\documentclass{article} 
\usepackage{iclr2025_conference,times}
\usepackage{colortbl}
\usepackage{xcolor}
\usepackage{subcaption}

\usepackage{amsmath,amsfonts,bm}

\def\eqref#1{equation~\ref{#1}}

\def\1{\bm{1}}

\DeclareMathAlphabet{\mathsfit}{\encodingdefault}{\sfdefault}{m}{sl}
\SetMathAlphabet{\mathsfit}{bold}{\encodingdefault}{\sfdefault}{bx}{n}

\newcommand{\R}{\mathbb{R}}

\usepackage[colorlinks]{hyperref}
\usepackage{url}
\usepackage{graphicx}

\usepackage{marvosym}

\title{Reflective Gaussian Splatting}

\author{Yuxuan Yao$^1$$^*$, Zixuan Zeng$^1$$^*$, Chun Gu$^1$, Xiatian Zhu$^{2\dag}$, Li Zhang$^{1\dag}$\textsuperscript{\Letter}
  \\
  $^1$School of Data Science, Fudan University 
  \quad
  $^2$University of Surrey 
  \vspace{.5em} 
  \\
  \url{https://fudan-zvg.github.io/ref-gaussian}
}

\newcommand{\model}{\textit{Ref-Gaussian}}

\iclrfinalcopy 
\begin{document}

\maketitle

\renewcommand{\thefootnote}{}
\footnotetext{$^*$Equally contributed}
\footnotetext{$^\dag$Co-last authorship}
\footnotetext{\textsuperscript{\Letter}Li Zhang (lizhangfd@fudan.edu.cn) is the corresponding author.}

\begin{figure}[ht]
    \centering
    \includegraphics[width=0.95\linewidth]{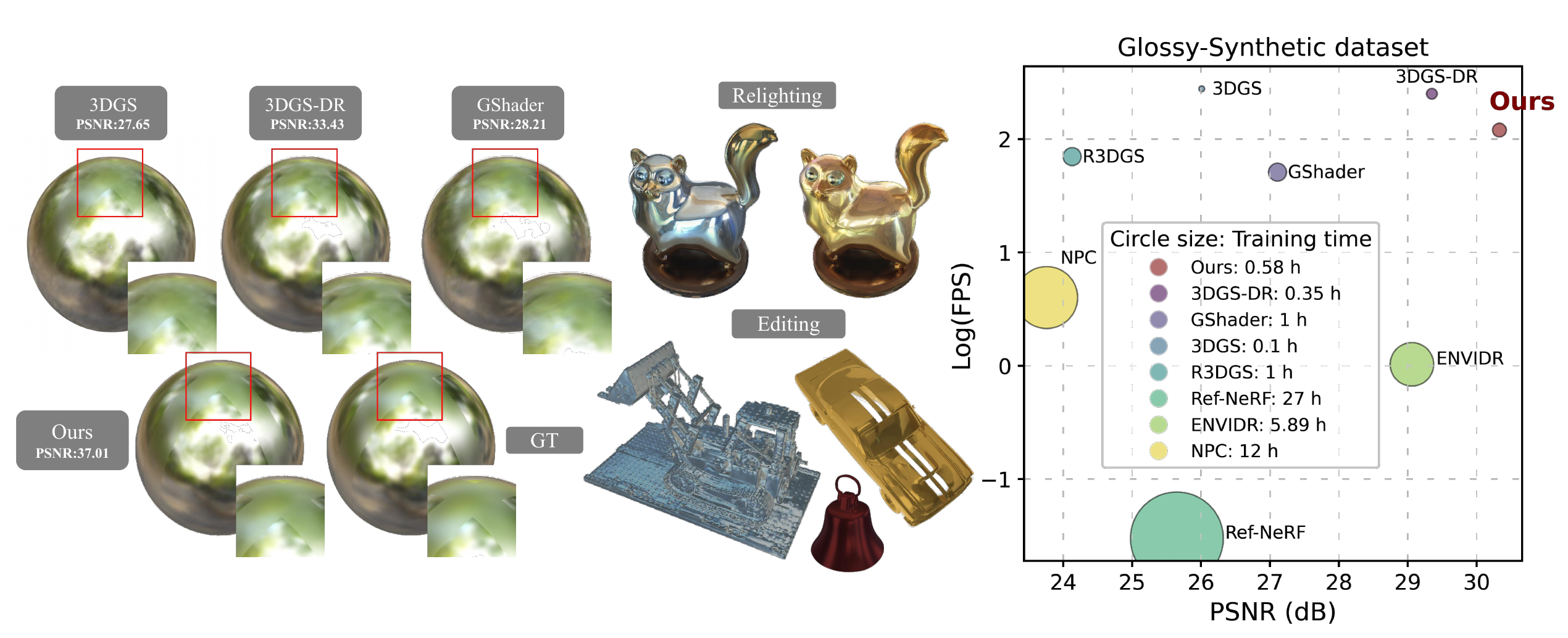}
    \caption{
    Our \model{} achieves superior rendering quality in novel view synthesis while enjoying fast optimization (training time) and real-time rendering (FPS), and 
    supporting various downstream applications such as \textit{relighting} and \textit{editing}.
    Please refer to the video in our \textit{supplementary material} for a more comprehensive and intuitive comparison.
    }
    \label{fig:teaser}
\end{figure}

\begin{abstract}

Novel view synthesis has experienced significant advancements owing to increasingly capable NeRF- and 3DGS-based methods. However, reflective object reconstruction remains challenging, lacking a proper solution to achieve real-time, high-quality rendering while accommodating inter-reflection. To fill this gap, we introduce a Reflective Gaussian splatting (\textbf{\model{}}) framework characterized with two components: (I) {\em Physically based deferred rendering} that empowers the rendering equation with pixel-level material properties via formulating split-sum approximation; (II) {\em Gaussian-grounded inter-reflection} that realizes the desired inter-reflection function within a Gaussian splatting paradigm for the first time. To enhance geometry modeling, we further introduce material-aware normal propagation and an initial per-Gaussian shading stage, along with 2D Gaussian primitives. Extensive experiments on standard datasets demonstrate that \model{} surpasses existing approaches in terms of quantitative metrics, visual quality, and compute efficiency. Further, we show that our method serves as a unified solution for both reflective and non-reflective scenes, going beyond the previous alternatives focusing on only reflective scenes. Also, we illustrate that \model{} supports more applications such as relighting and editing.

\end{abstract}

\section{Introduction}

Tremendous progress has been recently made in 3D object reconstruction and novel view synthesis from multiview images. Neural radiance field (NeRF)~\citep{mildenhall2021nerf} represents 3D scenes with neural implicit fields, achieving high-quality, photo-realistic renderings through volume rendering. However, NeRF is expensive computationally for optimization and rendering, limiting its use in real-time applications. To address this, 3D Gaussian splatting (3DGS)~\cite{kerbl20233d} instead represents a scene using a set of 3D Gaussians. Combining rasterization and alpha-blending, it achieves both real-time and high-quality rendering.

However, most NeRF and 3DGS struggle to model {\em reflective surfaces} 
due to the intrinsic inability of capturing high-frequency specular components.
To mitigate this, Ref-NeRF~\citep{verbin2022ref} 
leverages integrated directional encoding to smooth interpolation of reflected radiance. But this method fails to decompose the environment illumination,
making it less useful for example unable to relight. 
A couple of NeRF methods~\citep{verbin2022ref, liang2023envidr} are challenged by computational cost and limited flexibility.
Using a simplified shading functions on each Gaussian can achieve rapid convergence,
but suffering substantial noise in modeling of geometry, material, and lighting~\citep{jiang2024gaussianshader}.
To avoid this issue, performing the shading function at the pixel level (i.e., deferred shading) following alpha-blending to smooth out the gradients is helpful~\citep{ye20243d}.
Nonetheless, limiting to a simplified shading function makes it ineffective
to model complex reflective objects with complex structures and surfaces and the inter-reflection phenomenon. To address the latter, RelightableGaussian~\citep{gao2023relightable} traces rays across Gaussians for visibility inferring at the cost of heavy noises and added computational overhead from Monte Carlo sampling.

To address the aforementioned issues all together,
we propose a Reflective Gaussian splatting (\textbf{\model{}}) framework,
allowing for real-time high-quality rendering of reflective objects
while taking into account the complicated inter-reflection effect.
This is made possible by two key components:
\textbf{(I)} {\em Physically based deferred rendering}, where 
we empower the rendering equation with pixel-level material properties (e.g., Bidirectional reflectance distribution function, or BRDF in short),
and adopt the split-sum approximation to avoid the heavy compute with Monte Carlo sampling.
\textbf{(II)} {\em Gaussian-grounded inter-reflection}, 
where we realize the intricate inter-reflection function with Gaussian splatting
for the very first time, despite the challenge of extracting the mesh accurately.
To enhance the modeling of geometry,
we choose 2D Gaussian primitives as scene representation,
while introducing an initial per-Gaussian shading stage (performing shading at the Gaussian level to provide good geometric initialization),
as well as a material-aware normal propagation process (enlarging the scales of Gaussians with high metallic and low roughness).

We make these \textbf{contributions}:
\textbf{(I)} We strive to realize real-time high-quality rendering of
reflective objects with inter-reflection in 3D space.
\textbf{(II)} We innovate a generic 3D reconstruction framework,
Reflective Gaussian splatting (\textbf{\model{}}), characterized with
{\em physically based deferred rendering}
and {\em Gaussian-grounded inter-reflection}.
This approach is further enhanced by geometry-focused optimization including adopting 2D Gaussian primitives, material-aware normal propagation, and per-Gaussian shading initialization.
\textbf{(III)} Extensive experiments demonstrate that \model{} outperforms previous methods in terms of quantitative metrics, visual quality, and compute efficiency under both reflective and non-reflective scenes,
whilst supporting downstream applications such as relighting and editing.

\section{Related work}
\label{gen_inst}

\paragraph{Novel view synthesis} aims to generate new images of a scene from unseen viewpoints with a limited set of observed images.
Neural radiance fields (NeRF)~\citep{mildenhall2020nerf} employ volumetric rendering with implicit neural representations to produce highly detailed novel views from input images. Subsequent works have focused on two main directions: improving rendering quality~\citep{barron2021mip,barron2022mip,barron2023zipnerf} and accelerating both training and rendering speeds~\citep{SunSC22,yu_and_fridovichkeil2021plenoxels,muller2022instant,Chen2022ECCV}. More recently, 3D Gaussian splatting (3DGS)~\citep{kerbl20233d} proposed using 3D Gaussian primitives and tile-based differentiable rasterizer, achieving superior rendering quality and efficiency over NeRF-based methods. Building on 3DGS, methods like 2DGS~\citep{huang20242d} and GOF~\citep{yu2024gaussian} have further explored surface reconstruction with Gaussian splatting. In this paper, we adopt 2D Gaussian primitives, as proposed in 2DGS, to represent scenes for more accurate geometry reconstruction.

\paragraph{Inverse rendering} seeks to decompose scenes into geometry, lighting, and materials, posing a significant challenge due to the high-dimensional complexity of light interaction. NeRF~\citep{mildenhall2020nerf} is commonly used as a 3D representation in inverse rendering due to its flexible raymarching technique. Several NeRF-based methods~\citep{srinivasan2021nerv,boss2021nerd,zhang2021nerfactor,boss2021neural,yao2022neilf,boss2022samurai,zhang2023neilf++,liu2023nero,jin2023tensoir} model light interaction using the explicit rendering equation with BRDF. For instance, NeRO~\citep{liu2023nero} divides the training process into two stages: the first stage employs split-sum approximation to obtain detailed geometry, while the second stage focuses on accurate sampling to recover lighting and materials. In contrast, Ref-NeRF~\citep{verbin2022ref} does not decompose material and environment lighting but instead shades points using a simplified function, limiting its applicability in downstream tasks such as relighting and editing.

With the advent of 3D Gaussian splatting~\citep{kerbl20233d}, recent methods~\citep{jiang2024gaussianshader,gao2023relightable,liang2024gs,shi2023gir} have begun to address inverse rendering tasks using Gaussian primitives for much faster rendering speed than NeRF methods. GShader~\citep{jiang2024gaussianshader} applies a simplified shading functions from rendering equation on each Gaussian. While it achieves rapid convergence, this approach also brings substantial noise in geometry, material, and lighting. 
In contrast, 3DGS-DR~\citep{ye20243d} employs deferred shading by applying a simplified shading function at the pixel level, stabilizing the optimization process by smoothing out normal gradients. As both GShader and 3DGS-DR cannot model the inter-reflection effects, RelightableGaussian~\citep{gao2023relightable} introduces an innovative Gaussian-based ray tracing technique for visibility precomputation. However, it applies Monte Carlo sampling to evaluate the rendering equation, which slows down the rendering speed, and the inaccurate visibility also reduces the quality of reconstruction. In this paper, inspired by 3DGS-DR, we adopt pixel-level deferred shading but perform the rendering function with BRDF properties to handle complex scenes. Furthermore,  we propose performing ray tracing on the extracted mesh for visibility computation in the specular component of our rendering equation. Together with the split-sum approximation of the rendering equation,  we model inter-reflection effects while maintaining fast rendering speeds.

\section{Method}

\begin{figure}[t]
    \centering
    \includegraphics[width=\linewidth]{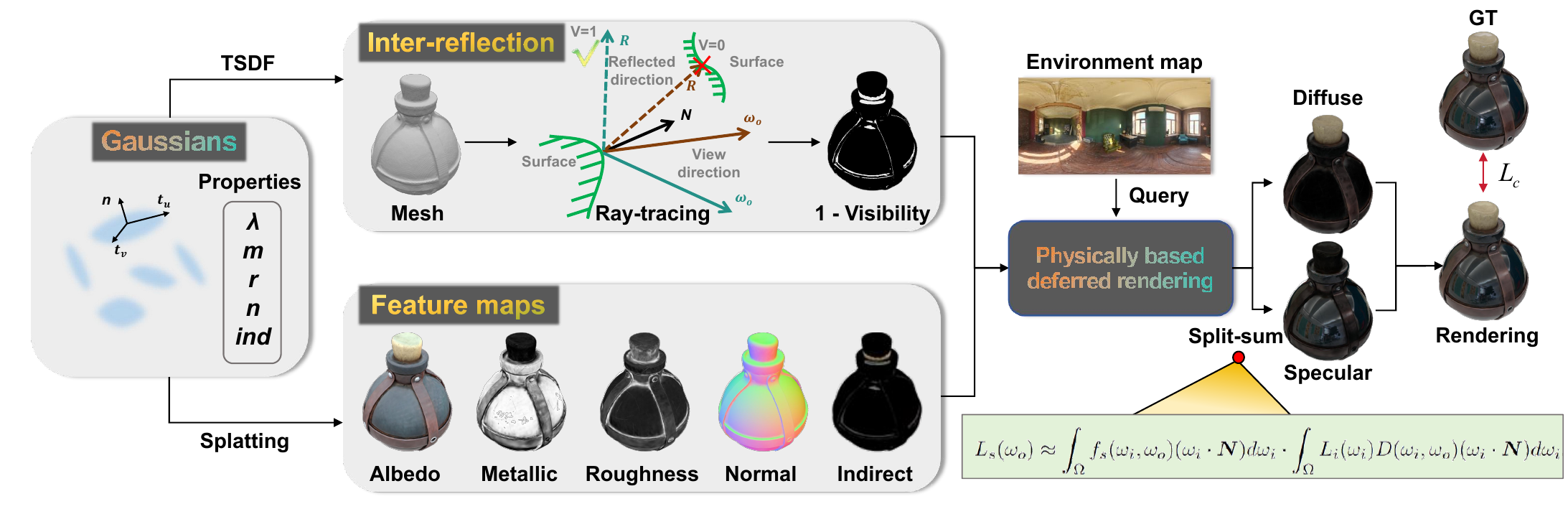}
    \caption{
    Overview of the \textbf{\model{}} framework: First, we apply the splatting process to produce feature maps and perform ray-tracing on the extracted mesh to compute visibility for the specular term in the rendering equation. Next, we use the pixel-level feature maps to apply the rendering equation with split-sum approximation, yielding the final physically based rendering result.
    }
    \label{fig:pipeline}
\end{figure}

Our Reflective Gaussian splatting (\model{}) is formulated in the Gaussian splatting framework. 
The key components of \model{} include (1) physically based deferred rendering and (2) Gaussian-focused inter-reflection.
An overview of our pipeline is shown in Figure~\ref{fig:pipeline}.
We start with the preliminary of Gaussian splatting.

In standard 3DGS~\citep{kerbl20233d}, each Gaussian is modeled as an ellipsoid, characterized by a 3D position $\boldsymbol{p}_k$ and a covariance matrix $\Sigma$, defined as:
\begin{equation}
\mathcal{G}(\boldsymbol{p}) = \exp \left( -\frac{1}{2} (\boldsymbol{p} - \boldsymbol{p}_k)^\mathrm{T} \Sigma^{-1} (\boldsymbol{p} - \boldsymbol{p}_k) \right).
\end{equation}
During the rendering process, 3D Gaussians are projected into 2D Gaussians in camera space. The transformed 2D covariance matrix is approximated as: $\Sigma' = JW\Sigma W^\mathrm{T} J^\mathrm{T}$, where $W$ is the view transformation matrix and $J$ is the Jacobian of the perspective projection. Each Gaussian is also assigned an opacity $o$ and a view-dependent color $\boldsymbol{c}_i$ represented by spherical harmonics. The final pixel color $\boldsymbol{C}$ is computed by blending Gaussians in front-to-back order based on their depth:
\begin{equation}
\label{eq:alphab-lending}
\boldsymbol{C} = \sum_{i=1}^N \boldsymbol{c}_i T_i \alpha_i, \quad \text{where} \quad T_i = \prod_{j=1}^{i-1} (1 - \alpha_j), 
\end{equation} 
where $\alpha_i$ is calculated by multiplying the opacity $o_i$ with the Gaussian weight, which is determined using the 2D covariance matrix $\Sigma'_i$.

\paragraph{2D Gaussian primitive}
However, 3D Gaussians struggle to capture detailed geometry due to inherent multi-view inconsistencies. 
We hence leverage 2D Gaussian splatting~\citep{huang20242d}
where view-consistent 2D oriented planar Gaussian disks are used as rendering primitives.

Each 2D Gaussian is defined by its position $\boldsymbol{p}$, tangential vectors $\boldsymbol{t}_u$ and $\boldsymbol{t}_v$, and scaling factors $(s_u, s_v)$ that control its shape and size. Its influence is given by:
\begin{equation}
\mathcal{G}(u,v) = \exp \left( -\frac{u^2+v^2}{2} \right),
\end{equation}
where $(u,v)$ are coordinates in the local tangent space. The ray-splat intersection $(u,v)$ in this space is obtained via:
\begin{equation}
    \boldsymbol{x} = (xz,yz,z,1)^\mathrm{T} = W H (u,v,1,1)^\mathrm{T}, \quad
    \text{where} \, H = 
    \begin{bmatrix}
        s_u \boldsymbol{t_u} & s_v \boldsymbol{t_v} & \boldsymbol{0} & \boldsymbol{p} \\
        0 & 0 & 0 & 1 \\
    \end{bmatrix} \in \R^{4\times4}.
    \label{eq:plane-to-world}
\end{equation}

Here, $\boldsymbol{x}$ represents the homogeneous ray passing through pixel $(x,y)$ and intersecting the 2D Gaussian disk at depth $z$. 
See more details of 2DGS in~\citep{huang20242d}.

\subsection{Physically based deferred rendering}
\label{sec:pbr}

Our physically based deferred rendering employs a simplified version of the Disney BRDF model~\citep{burley2012physically}.
Specifically, each Gaussian is associated with a set of material-related properties, including albedo $\boldsymbol{\lambda}\in[0,1]^3$, metallic $m\in[0,1]$, and roughness $r\in[0,1]$. The normal vector $\boldsymbol{n}\in[0,1]^3$ can be derived from the tangential vectors of each 2D Gaussian using $\boldsymbol{n} = \boldsymbol{t_u} \times \boldsymbol{t_v}$. 
We deploy the rendering equation on pixel-level feature maps obtained via alpha-blending:
\begin{equation}\label{eq:extra rasterization}
    \boldsymbol{X} = \sum_{i=1}^N \boldsymbol{x}_i \alpha_i \prod_{j=1}^{i-1} (1 - \alpha_j), \quad \text{where} \quad \boldsymbol{X} = \begin{bmatrix} \boldsymbol{\Lambda} \\ M \\ R \\ \boldsymbol{N} \end{bmatrix}, \quad
\boldsymbol{x}_i = \begin{bmatrix} \lambda_i \\ m_i \\ r_i \\ \boldsymbol{n}_i \end{bmatrix}.
\end{equation}
Note, unlike shading directly on the Gaussians before alpha-blending~\citep{jiang2024gaussianshader}, our deferred shading treats alpha-blending as a smoothing filter.
That allows to stabilize the optimization of features and finally produce more cohesive rendering results.

With the aggregated material maps, including albedo $\boldsymbol{\Lambda}$, metallic $M$, roughness $R$, and normal $\boldsymbol{N}$, the rendering equation is subsequently applied. The rendering equation expresses the outgoing radiance $L(\omega_o)$ in the direction $\omega_o$ as:
\begin{equation}
L(\omega_o) = \int_{\Omega} L_i(\omega_i) f(\omega_i,\omega_o) (\omega_i \cdot \boldsymbol{N}) \, d\omega_i,
\end{equation}
where $f(\omega_i,\omega_o)$ denotes the bidirectional reflectance distribution function (BRDF). It represents the integral of reflected incident light over the hemisphere.

Since BRDF consists of both diffuse and specular terms, the integral can also be divided into two parts. The diffuse term is computed by querying the pre-integrated environment map using the normal $\boldsymbol{N}$ and multiplying with the material terms, while the specular term requires more complex reflection calculations to account for surface reflections. Here, let us focus on the specular component, where the specular term of the BRDF is given by: 
\begin{equation}
f_s(\omega_i,\omega_o) = \frac{D \ G \ F }{4 (\omega_o \cdot \boldsymbol{N}) (\omega_i \cdot \boldsymbol{N})},
\end{equation}
where $D$, $F$, and $G$ represent the GGX normal distribution function, the Fresnel term, and the shadowing-masking term, respectively. 

To compute the above integral term, a typical approach is Monte Carlo sampling
as in~\citep{gao2023relightable}.
However, that is computationally expensive to render in real time.
Instead, we adopt the split-sum approximation~\citep{munkberg2022extracting}:

\begin{equation}\label{eq:split sum}
L_s(\omega_o) \approx \int_{\Omega} f_s(\omega_i, \omega_o) (\omega_i \cdot \boldsymbol{N}) d\omega_i \cdot \int_{\Omega} L_i(\omega_i) D(\omega_i, \omega_o) (\omega_i \cdot \boldsymbol{N}) d\omega_i,
\end{equation}
where the first term depends solely on $(\omega_i \cdot \boldsymbol{N})$ and roughness $R$, allowing the results to be precomputed and stored in a 2D lookup texture map. The second term represents the integral of incident radiance over the specular lobe, and can be pre-integrated at the beginning of each training iteration with different roughness levels. That allows to efficiently employ a series of cubemaps to represent the environment lighting by performing trilinear interpolation using the reflected direction and the roughness as parameters.

\subsection{Gaussian grounded inter-reflection}\label{sec:interreflection}
Inter-reflection plays a critical role in rendering reflective objects.
To that end, we further enhance the specular light component by separately modeling the direct light $L_\mathrm{dir}$ (second term in Eq.~\ref{eq:split sum}) and indirect light $L_\mathrm{ind}$. Direct light $L_\mathrm{dir}$ denotes that whose reflection is not blocked by any scene elements, and the rest is defined as indirect light $L_\mathrm{ind}$.

Specifically, we approximate the visibility $V \in \{0, 1\}$ of incident light based on whether it is self-occluded along the reflected direction $\boldsymbol{R}=2(w_o\cdot \boldsymbol{N})\boldsymbol{N}-w_o$. We represent the indirect light from occluded part as $L_\mathrm{ind}$: 
\begin{equation}\label{eq:indirect}
L_s'(\omega_o) \approx ( \int_{\Omega} f_s(\omega_i, \omega_o) (\omega_i \cdot \boldsymbol{N}) d\omega_i ) \cdot [L_\mathrm{dir} \cdot V + L_\mathrm{ind} \cdot (1 - V)]. 
\end{equation}
Intuitively, this indirect light component $L_\mathrm{ind}$ introduced here aims to effectively model the perturbation caused by occlusion in environmental lighting estimation.

Efficiently computing visibility along a ray across Gaussian primitives is non-trivial. Here, we propose ray tracing on extracted mesh on the fly. During optimization, we periodically extract the object's surface mesh using truncated signed distance function (TSDF) fusion. Once the mesh is constructed, we calculate the intersection points between rays and the surface using ray tracing, determining whether each pixel is occluded. For efficiency, we employ a bounding volume hierarchy (BVH) to accelerate the ray tracing process for visibility checks. As for direct lighting $ L_\mathrm{dir} $, it just  corresponds to the second term in Eq.\ref{eq:split sum}, handled as described in the previous section.

For the indirect lighting component, each Gaussian is assigned an additional view-dependent color $\boldsymbol{l}_\mathrm{ind}$, modeled by spherical harmonics. During the rendering process, $\boldsymbol{l}_\mathrm{ind}$ is evaluated in the reflected direction at the Gaussian level, and alpha blending is applied to aggregate the indirect lighting map as follows:
\begin{equation}
    L_\mathrm{ind} = \sum_{i=1}^N
    \boldsymbol{l}_\mathrm{ind} \alpha_i\prod_{j=1}^{i-1} (1 - \alpha_j).
\end{equation}

\subsection{Geometry focused model optimization}\label{sec:geometry}

To facilitate our physically based rendering,
we further enhance the underlying geometry during optimization
with the following designs.

\paragraph{Initial stage with per-Gaussian shading}
As the cost for the benefit of smoothing gradients,
pixel-level shading poses convergence challenges during the initial phase,
because the shading function disturbs the gradient directions of the geometry.
To address this, we propose a dedicated initial stage where the rendering equation is applied directly to each Gaussian primitive, using the material and geometry properties associated with it to compute the outgoing radiance. The outgoing radiance is then alpha-blended during rasterization to produce the final PBR rendering. This Gaussian-level shading design can help the gradients to be more effectively transferred back to the Gaussian primitives and facilitate eventually the optimization process.

\paragraph{Material-aware normal propagation}
Positions with inaccurate normals often have difficulty capturing a significant specular component due to its intrinsic sensitivity to the reflected direction. We observe that in \model{} there exists a strong positive correlation between normal accuracy and high metallic, low roughness properties~(a typical situation where the specular component is significant).
Under this insight, we propose periodically increasing the scale of 2D Gaussians with high metallic and low roughness so that their normal information can be propagated to adjacent Gaussians for capturing more accurate geometry.  

\paragraph{Objective loss function}
We design a geometry focused objective function:
$L = L_c + \lambda_n L_n + \lambda_\mathrm{smooth} L_\mathrm{smooth}$,
where $L_c = (1 - \lambda) L_1 + \lambda L_{\text{D-SSIM}}$ is the RGB reconstruction loss where we set the balancing weight $\lambda=0.2$~\citep{kerbl20233d}.

The second term $L_n = 1 - \tilde{\boldsymbol{N}}^\mathrm{T} \boldsymbol{N}$ is
the normal consistency loss. It encourages alignment of the Gaussians with the surface by minimizing the cosine difference between the rendered normal $\boldsymbol{N}$, obtained via alpha-blending, and the surface normal $\tilde{\boldsymbol{N}}$, derived from the depth map.

The third term $L_\mathrm{smooth} = \Vert \nabla \boldsymbol{N} \Vert \exp (-\Vert \nabla C_{gt} \Vert )$ is an edge-aware normal smoothness loss.
It aims to regularize normal variation in regions of low texture.

\begin{table}[t]
\caption{Per-scene image quality comparison on synthesized test views. 
The intensity of the red color signifies a better result.}
\label{tb:main_compare}
\vspace{0.3cm} 
\centering
\renewcommand{\arraystretch}{1.3} 
\resizebox{\textwidth}{!}{%
\begin{tabular}{l|cccccc|cccccccc|ccc}
\hline
\hline
\textbf{Datasets} & \multicolumn{6}{c|}{\textbf{Shiny Blender \citep{verbin2022ref}}} & \multicolumn{8}{c|}{\textbf{Glossy Synthetic \citep{liu2023nero}}} & \multicolumn{3}{c}{\textbf{Real~\citep{verbin2022ref}}} \\
\hline
\textbf{Scenes} & \textbf{ball} & \textbf{car} & \textbf{coffee} & \textbf{helmet} & \textbf{teapot} & \textbf{toaster} & \textbf{angel} & \textbf{bell} & \textbf{cat} & \textbf{horse} & \textbf{luyu} & \textbf{potion} & \textbf{tbell} & \textbf{teapot} & \textbf{garden} & \textbf{sedan} & \textbf{toycar} \\
\hline
\multicolumn{18}{c}{\textbf{PSNR} $\uparrow$} \\
\hline
Ref-NeRF & 33.16 & 30.44 & \cellcolor{red!10}33.99 & 29.94 & 45.12 & 26.12 & 20.89 & 30.02 & 29.76 & 19.30 & 25.42 & 30.11 & 26.91 & 22.77 & 22.01 & 25.21 & 23.65 \\
ENVIDR & \cellcolor{red!50}41.02 & 27.81 & 30.57 & \cellcolor{red!50}32.71 & 42.62 & 26.03 & 29.02 & \cellcolor{red!10}30.88 & 31.04 & 25.99 & 28.03 & 32.11 & 28.64 & \cellcolor{red!50}26.77 & 21.47 & 24.61 & 22.92 \\
3DGS & 27.65 & 27.26 & 32.30 & 28.22 & 45.71 & 20.99 & 24.49 & 25.11 & 31.36 & 24.63 & 26.97 & 30.16 & 23.88 & 21.51 & 21.75 & 26.03 & 23.78 \\
2DGS & 25.97 & 26.38 & 32.31 & 27.42 & 44.97 & 20.42 & 26.95 & 24.79 & 30.65 & 25.18 & 26.89 & 29.50 & 23.28 & 21.29 & \cellcolor{red!10}22.53 & 26.23 & 23.70 \\
GShader & 30.99 & 27.96 & 32.39 & 28.32 & 45.86 & 26.28 & 25.08 & 28.07 & 31.81 & \cellcolor{red!10}26.56 & 27.18 & 30.09 & 24.48 & 23.58 & 21.74 & 24.89 & 23.76 \\
R3DG & 23.64 & 25.92 & 30.10 & 25.01 & 43.15 & 18.80 & 24.90 & 23.51 & 27.59 & 23.37 & 24.68 & 27.29 & 21.25 & 20.47 & 21.92 & 21.18 & 22.83 \\
3DGS-DR & 33.43 & \cellcolor{red!10}30.48 & \cellcolor{red!10}34.53 & 31.44 & \cellcolor{red!10}47.04 & \cellcolor{red!10}26.76 & \cellcolor{red!10}29.07 & 30.60 & \cellcolor{red!10}32.59 & 26.17 & \cellcolor{red!10}28.96 & \cellcolor{red!10}32.65 & \cellcolor{red!10}29.03 & 25.77 & 21.82 & \cellcolor{red!10}26.32 & \cellcolor{red!10}23.83 \\
\textbf{\model{}} & \cellcolor{red!10}37.01 & \cellcolor{red!50}31.04 & \cellcolor{red!50}34.63 & \cellcolor{red!10}32.32 & \cellcolor{red!50}47.16 & \cellcolor{red!50}28.05 & \cellcolor{red!50}30.38 & \cellcolor{red!50}32.86 & \cellcolor{red!50}33.01 & \cellcolor{red!50}27.05 & \cellcolor{red!50}30.04 & \cellcolor{red!50}33.07 & \cellcolor{red!50}29.84 & \cellcolor{red!10}26.68 & \cellcolor{red!50}22.97 & \cellcolor{red!50}26.60 & \cellcolor{red!50}24.27 \\
\hline
\multicolumn{18}{c}{\textbf{SSIM} $\uparrow$} \\
\hline
Ref-NeRF & 0.971 & 0.950 & \cellcolor{red!10}0.972 & 0.954 & 0.995 & 0.921 & 0.853 & 0.941 & 0.944 & 0.820 & 0.901 & 0.933 & 0.947 & 0.897 & \cellcolor{red!10}0.584 & 0.720 & 0.633 \\
ENVIDR & \cellcolor{red!50}0.997 & 0.943 & 0.962 & \cellcolor{red!50}0.987 & 0.995 & 0.922 & 0.934 & 0.954 & \cellcolor{red!10}0.965 & 0.925 & 0.931 & \cellcolor{red!10}0.960 & 0.947 & \cellcolor{red!50}0.957 & 0.561 & 0.707 & 0.549 \\
3DGS & 0.937 & 0.931 & \cellcolor{red!10}0.972 & 0.951 & 0.996 & 0.894 & 0.792 & 0.908 & 0.959 & 0.797 & 0.916 & 0.938 & 0.900 & 0.881 & 0.571 & 0.771 & 0.637 \\
2DGS & 0.934 & 0.930 & \cellcolor{red!10}0.972 & 0.953 & 0.997 & 0.892 & 0.918 & 0.911 & 0.958 & 0.909 & 0.918 & 0.939 & 0.902 & 0.886 & 0.609 & 0.778 & 0.597 \\
GShader & 0.966 & 0.932 & 0.971 & 0.951 & 0.996 & 0.929 & 0.914 & 0.919 & 0.961 & 0.933 & 0.914 & 0.936 & 0.898 & 0.901 & 0.576 & 0.728 & 0.637 \\
R3DG & 0.888 & 0.922 & 0.963 & 0.931 & 0.995 & 0.858 & 0.894 & 0.888 & 0.934 & 0.878 & 0.889 & 0.911 & 0.875 & 0.869 & 0.556 & 0.643 & \cellcolor{red!10}0.657 \\
3DGS-DR & 0.979 & \cellcolor{red!10}0.963 & \cellcolor{red!50}0.976 & 0.971 & \cellcolor{red!10}0.997 & \cellcolor{red!10}0.942 & \cellcolor{red!10}0.942 & \cellcolor{red!10}0.959 & \cellcolor{red!50}0.973 & \cellcolor{red!10}0.933 & \cellcolor{red!10}0.943 & 0.959 & \cellcolor{red!10}0.958 & 0.942 & 0.581 & \cellcolor{red!10}0.773 & 0.639 \\
\textbf{\model{}} & \cellcolor{red!10}0.981 & \cellcolor{red!50}0.964 & \cellcolor{red!50}0.976 & \cellcolor{red!10}0.971 & \cellcolor{red!50}0.998 & \cellcolor{red!50}0.948 & \cellcolor{red!50}0.954 & \cellcolor{red!50}0.969 & \cellcolor{red!50}0.973 & \cellcolor{red!50}0.944 & \cellcolor{red!50}0.952 & \cellcolor{red!50}0.963 & \cellcolor{red!50}0.962 & \cellcolor{red!10}0.947 & \cellcolor{red!50}0.617 & \cellcolor{red!50}0.777 & \cellcolor{red!50}0.660 \\
\hline
\multicolumn{18}{c}{\textbf{LPIPS} $\downarrow$} \\
\hline
Ref-NeRF & 0.166 & 0.050 & 0.082 & 0.086 & 0.012 & 0.083 & 0.144 & 0.102 & 0.104 & 0.155 & 0.098 & 0.084 & 0.114 & 0.098 & 0.251 & 0.234 & \cellcolor{red!50}0.231 \\
ENVIDR & \cellcolor{red!50}0.020 & 0.046 & 0.083 & \cellcolor{red!50}0.036 & 0.009 & \cellcolor{red!10}0.081 & 0.067 & 0.054 & 0.049 & 0.065 & 0.059 & 0.072 & 0.069 & \cellcolor{red!50}0.041 & 0.263 & 0.387 & 0.345 \\
3DGS & 0.162 & 0.047 & 0.079 & 0.081 & 0.008 & 0.125 & 0.088 & 0.104 & 0.062 & 0.077 & 0.064 & 0.093 & 0.102 & 0.125 & \cellcolor{red!10}0.248 & \cellcolor{red!10}0.206 & \cellcolor{red!10}0.237 \\
2DGS & 0.156 & 0.052 & 0.079 & 0.079 & 0.008 & 0.127 & 0.072 & 0.109 & 0.060 & 0.071 & 0.066 & 0.097 & 0.125 & 0.101 & 0.254 & 0.225 & 0.396 \\
GShader & 0.121 & \cellcolor{red!10}0.044 & \cellcolor{red!10}0.078 & 0.074 & \cellcolor{red!10}0.007 & 0.079 & 0.082 & 0.098 & 0.056 & 0.562 & 0.064 & 0.088 & 0.091 & 0.122 & 0.274 & 0.259 & 0.239 \\
R3DG & 0.214 & 0.058 & 0.090 & 0.125 & 0.013 & 0.170 & 0.085 & 0.125 & 0.089 & 0.081 & 0.080 & 0.117 & 0.156 & 0.115 & 0.354 & 0.380 & 0.312 \\
3DGS-DR & 0.105 & \cellcolor{red!50}0.033 & \cellcolor{red!50}0.076 & 0.050 & \cellcolor{red!50}0.006 & 0.082 & \cellcolor{red!10}0.052 & \cellcolor{red!10}0.050 & \cellcolor{red!10}0.042 & \cellcolor{red!10}0.057 & \cellcolor{red!10}0.048 & \cellcolor{red!10}0.068 & \cellcolor{red!10}0.059 & 0.060 & \cellcolor{red!50}0.247 & \cellcolor{red!50}0.208 & \cellcolor{red!50}0.231 \\
\textbf{\model{}} & \cellcolor{red!10}0.098 & \cellcolor{red!50}0.033 & \cellcolor{red!50}0.076 & \cellcolor{red!10}0.049 & \cellcolor{red!50}0.006 & \cellcolor{red!50}0.074 & \cellcolor{red!50}0.042 & \cellcolor{red!50}0.040 & \cellcolor{red!50}0.040 & \cellcolor{red!50}0.048 & \cellcolor{red!50}0.043 & \cellcolor{red!50}0.064 & \cellcolor{red!50}0.058 & \cellcolor{red!10}0.058 & 0.256 & 0.245 & 0.256\\
\hline
\hline
\end{tabular}%
}
\end{table}

\section{Experiments}
To evaluate the performance of \model{} in novel view synthesis, as well as geometry, materials, and lighting reconstruction, and downstream applications, we conduct a series of quantitative and qualitative experiments. Additionally, We provide a demo video in the \textit{supplementary material}, showcasing dynamic renderings of \model{} across two synthetic datasets, along with a comparative analysis.

{\bf Datasets}
We select two synthetic datasets Shiny Blender~\citep{verbin2022ref} and Glossy Synthetic~\citep{liu2023nero} for novel view synthesis of reflective objects, and Ref-Real dataset~\citep{verbin2022ref} for real-world open scenes. 

{\bf Competitors} We compare several representative models in the field of reflective 3D reconstruction, including Ref-NeRF~\citep{verbin2022ref}, ENVIDR~\citep{liang2023envidr}, 3DGS~\citep{kerbl20233d}, GaussianShader~\citep{jiang2024gaussianshader}, RelightableGaussian (R3DG)~\citep{gao2023relightable}, and 3DGS-DR~\citep{ye20243d}. We also compared with more NeRF-based models like NeRO~\citep{liu2023nero}, NeuS~\citep{wang2021neus} and NDE~\citep{wu2024neural}, see Tables~\ref{tb:neus_nero} and~\ref{tb:nde} in appendix.

{\bf Evaluation metrics} We use three standard metrics: PSNR, SSIM~\citep{wang2004image}, and LPIPS~\citep{zhang2018unreasonable}. 

\begin{figure}[t]
    \centering
    \includegraphics[width=\linewidth]{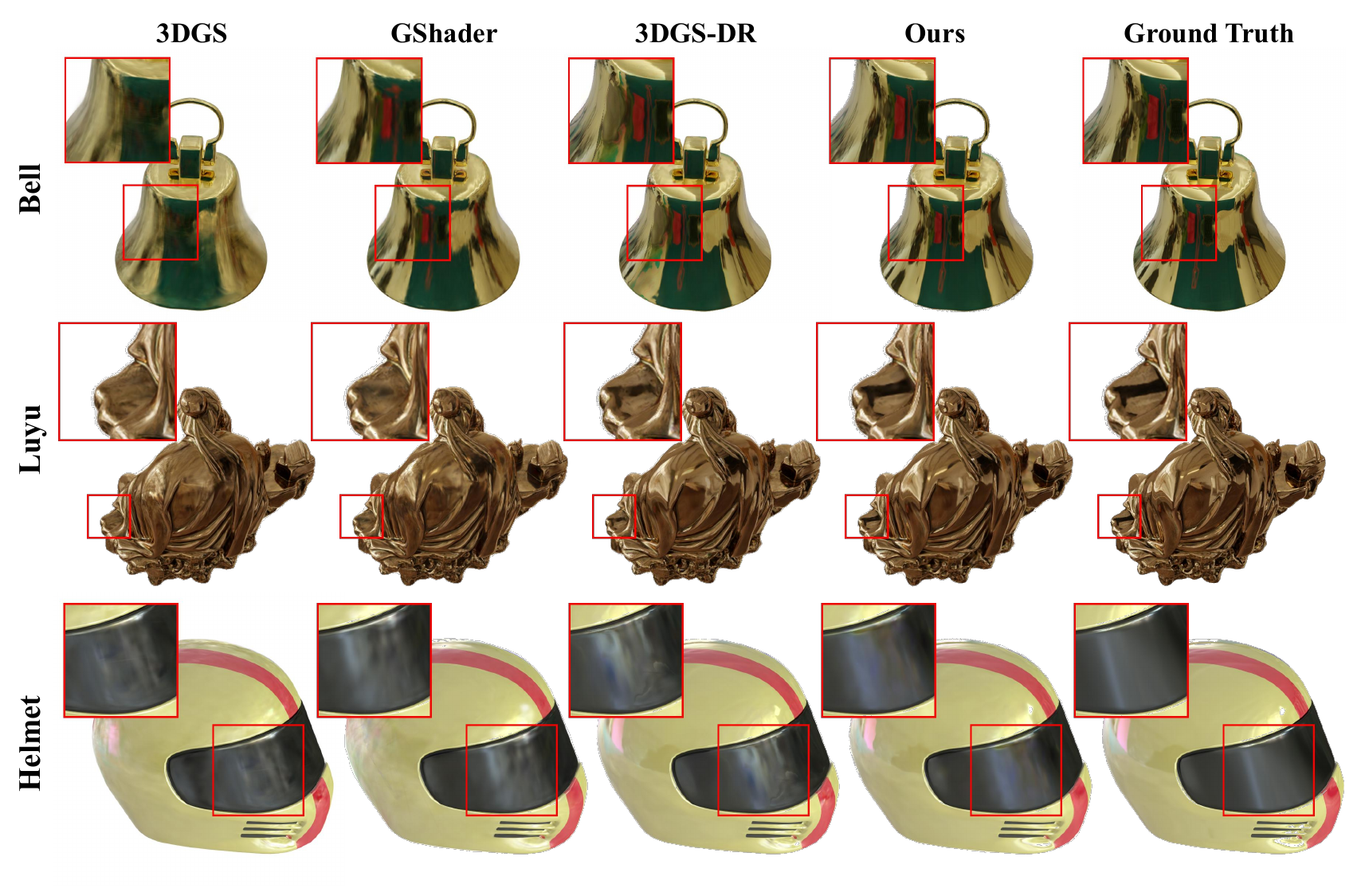}
    \caption{
        Qualitative comparisons on  reflective scenes, including Bell, Luyu and Helmet. 
    }
    \label{fig:compare-main}
\end{figure}
{\bf Implementation details} Our training procedure consists of two stages. We utilize a per-Gaussian rendering for $18,000$ steps as an initial stage, followed by a deferred rendering stage training for about $40,000$ steps. We reset all color and material attributes before the second stage, retaining only the the geometry of Gaussians. The learning rates of the trainable material attributes (metallic, roughness, and albedo) are all set to $0.005$, while the learning rate of the environment map is set to $0.01$. The set of other basic trainable attributes for Gaussians like position and covariance is consistent with 2DGS~\citep{huang20242d}. Then, our material-aware normal propagation is conducted to those Gaussians with metallic no less than $0.02$ and roughness no more than $0.1$. Also, we adopt the metallic initial value from 3DGS-DR~\citep{ye20243d} and set the initial roughness value to be $0.1$. During the implementation of physically based rendering, we discovered that spherical harmonics have a better fitting capability than the integrated diffuse lighting and we thereby use spherical harmonics to substitute it. For the loss function, $\lambda_n$ is set to $0.05$ and $\lambda_\mathrm{smooth}$ is $1.0$. Additionally, we periodically extract object's surface mesh at 3000 step intervals. All experiments are conducted on a single NVIDIA A6000 GPU.

\begin{figure}[!t]
    \centering
    \includegraphics[width=\linewidth]{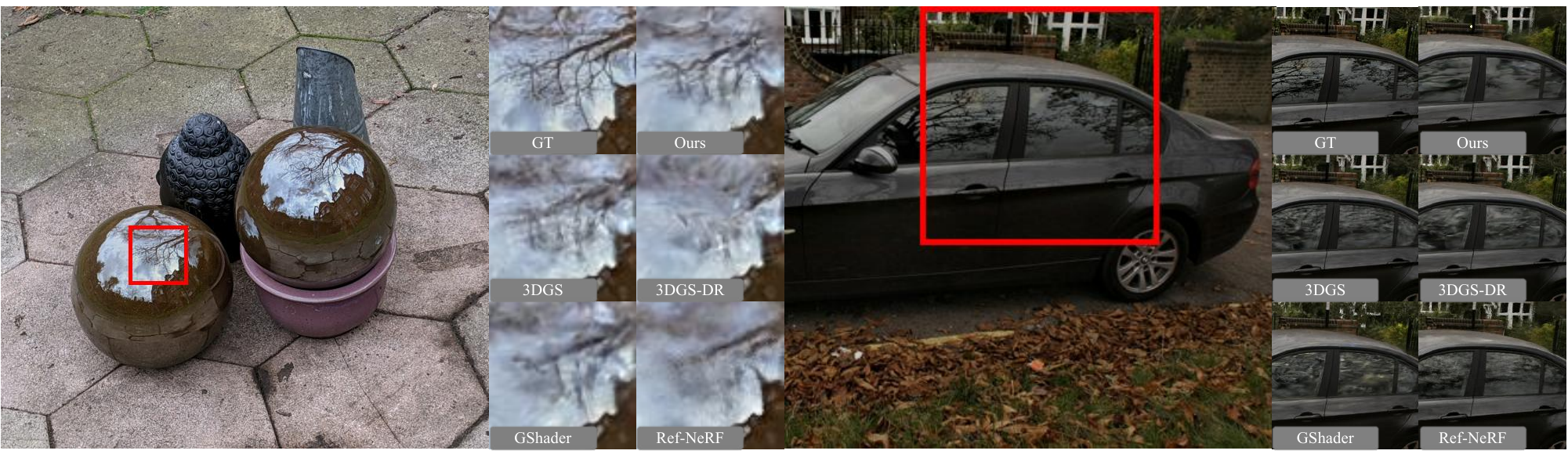}
    \caption{
        Qualitative comparisons on Ref-Real dataset~\citep{verbin2022ref}.
    }
    \label{fig:compare-real}
\end{figure}

\begin{figure}[t]
    \centering
    \includegraphics[width=\linewidth]{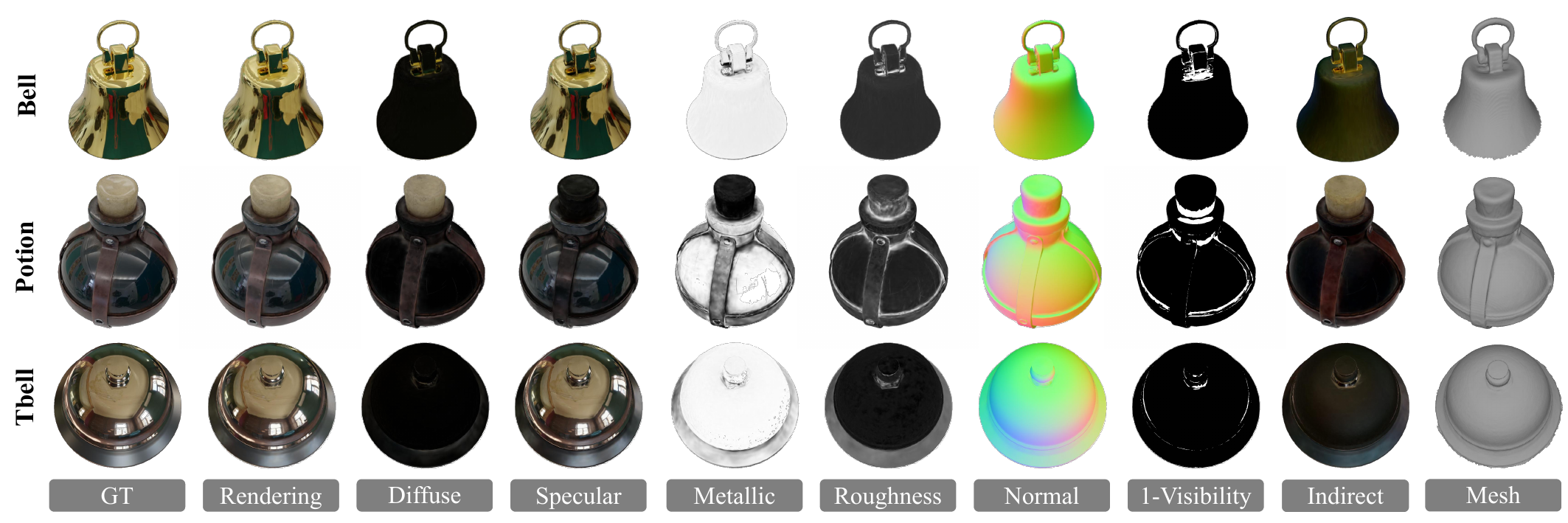}
    \caption{
        Inverse rendering with extracted mesh and indirect light. \textbf{Indirect}: Only consider indirect light as specular component when rendering.
    }
    \label{fig:inverse_mesh}
\end{figure} 
\begin{figure}[!t]
    \centering
    \includegraphics[width=\linewidth]{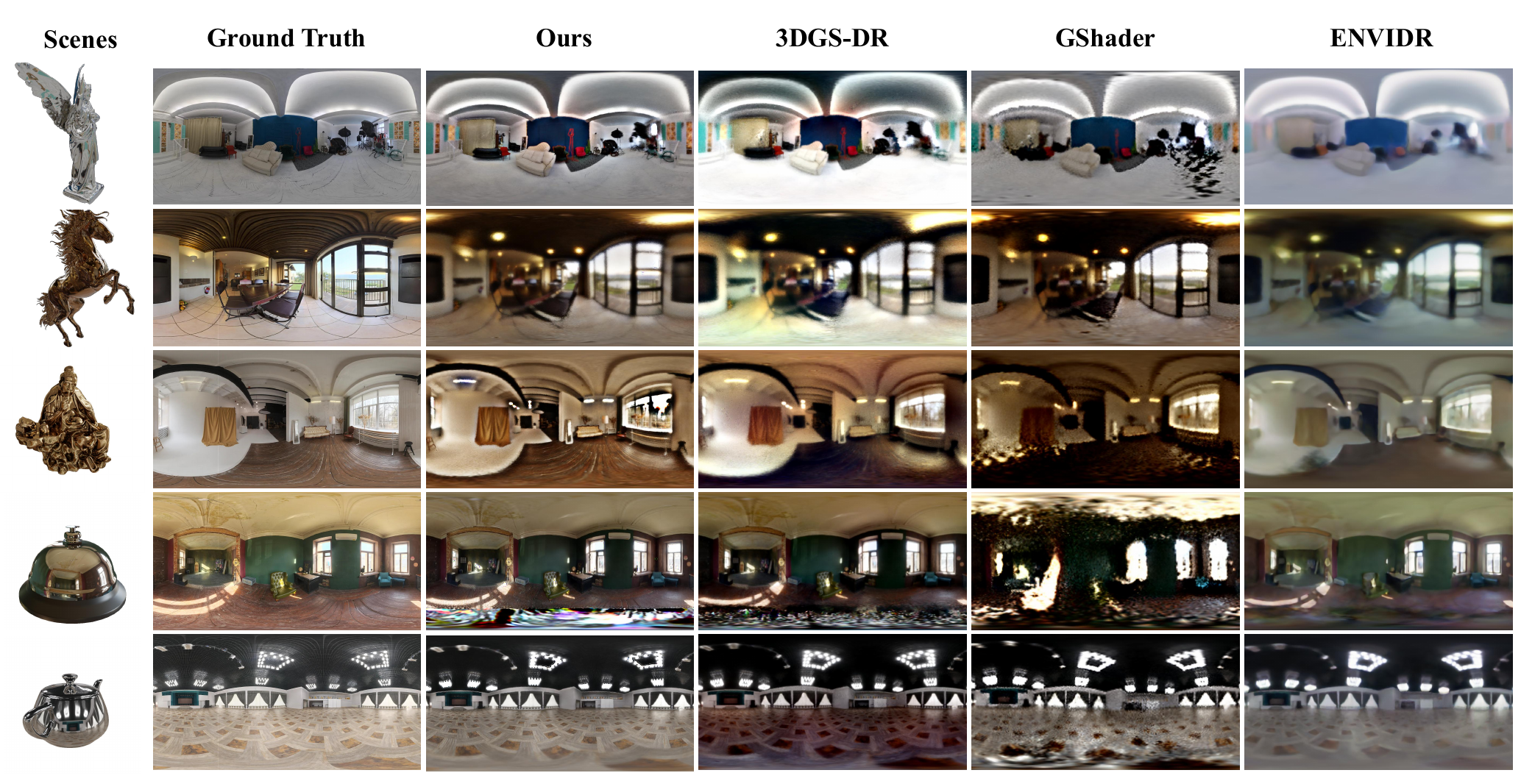}
    \caption{
        Qualitative comparisons of the estimated environment maps on Glossy Synthetic dataset~\citep{liu2023nero}.
    }
    \label{fig:compare-envmap}
\end{figure}

\subsection{Comparisons}

\paragraph{Novel view synthesis}
In Table \ref{tb:main_compare}, we present the quantitative results of several competitors, along with \model{}, for the novel view synthesis task across multiple datasets. \model{} achieves state-of-the-art performance in most scenes, especially with significant improvements on glossy synthetic datasets, where reflective surfaces dominate. As shown in Figure \ref{fig:compare-main}, 3DGS~\citep{kerbl20233d} struggles to handle high-frequency, view-dependent effects, resulting in noticeable blurriness on reflective surfaces in multi-view scenarios, such as Bell. GShader~\citep{jiang2024gaussianshader} introduces significant noise, leading to distortion. 3DGS-DR~\citep{ye20243d}, which employs a simplified shading function, finds it challenging to accurately model complex reflective scenes, as observed in Luyu and the rough tinted visor on Helmet. In contrast, \model{} effectively overcomes these challenges, offering precise environmental lighting modeling and robust handling of complex geometries and materials. 
Additionally, in Figure~\ref{fig:compare-real}, we present a qualitative comparison of real-world scenes~\citep{verbin2022ref}. \model{} provides more accurate normal maps and lighting, capturing subtle nuances on reflective surfaces. Specifically, the reflection phenomena on the garden sphere in the Garden and the car window in the Sedan are particularly noticeable. We demonstrate the more physically accurate reflection detail capture ability of \model{} in Figure~\ref{fig:compare-real}. 



\paragraph{Material, normal, and light estimation}
In Table~\ref{tb:Comprehensive Comparison}, we present a quantitative comparison against competitors in terms of normal maps, environment maps, and efficiency. \model{} demonstrates superior geometry reconstruction, as demonstrated by the high quality of the normal maps. This improvement is largely attributed to the carefully designed geometry optimization strategies discussed in Section~\ref{sec:geometry}. As shown in Figure~\ref{fig:inverse_mesh}, our normal map is smooth, as well as other decomposed material maps. Additionally, our visibility map is sharp and accurately highlights areas occluded in the reflected direction. \model{} also achieves the best results among competitors in estimating environment lighting, which is evaluated after re-scaling to eliminate ambiguity. 
Figure~\ref{fig:compare-envmap} presents a qualitative comparison of the estimated environment maps. Compared to previous methods, our results more closely resemble the ground truth in terms of both detail and overall tone, while exhibiting the least amount of noise. This improvement can be attributed to the precise geometry and comprehensive modeling employed in \model{}.

\begin{figure}[!t]
    \centering
    \begin{minipage}{0.37\linewidth}
        \centering
        \includegraphics[width=\linewidth]{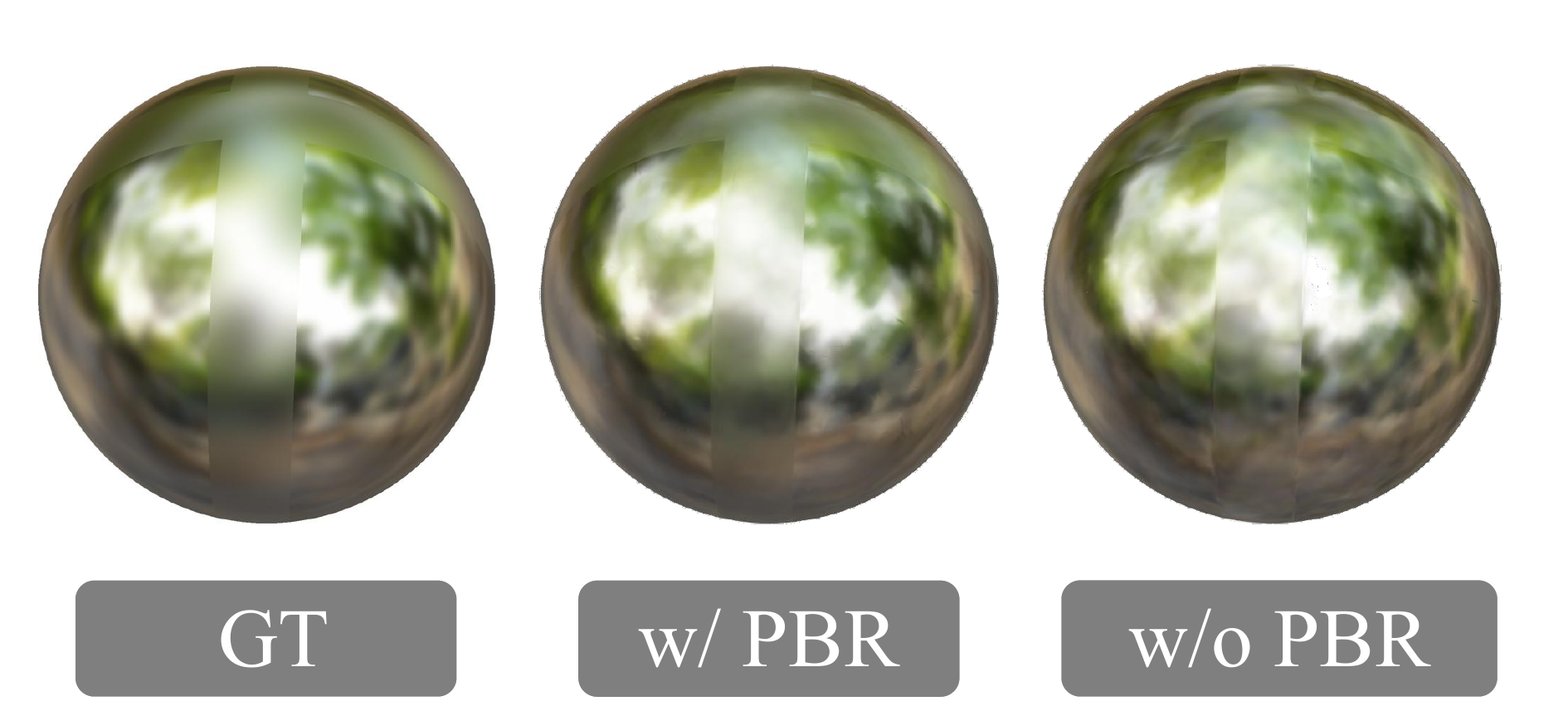}
        \caption{Ablation study on PBR.}
        \label{fig:PBR}
    \end{minipage}
    \hspace{0.02\linewidth}
    \begin{minipage}{0.59\linewidth}
        \centering
        \includegraphics[width=\linewidth]{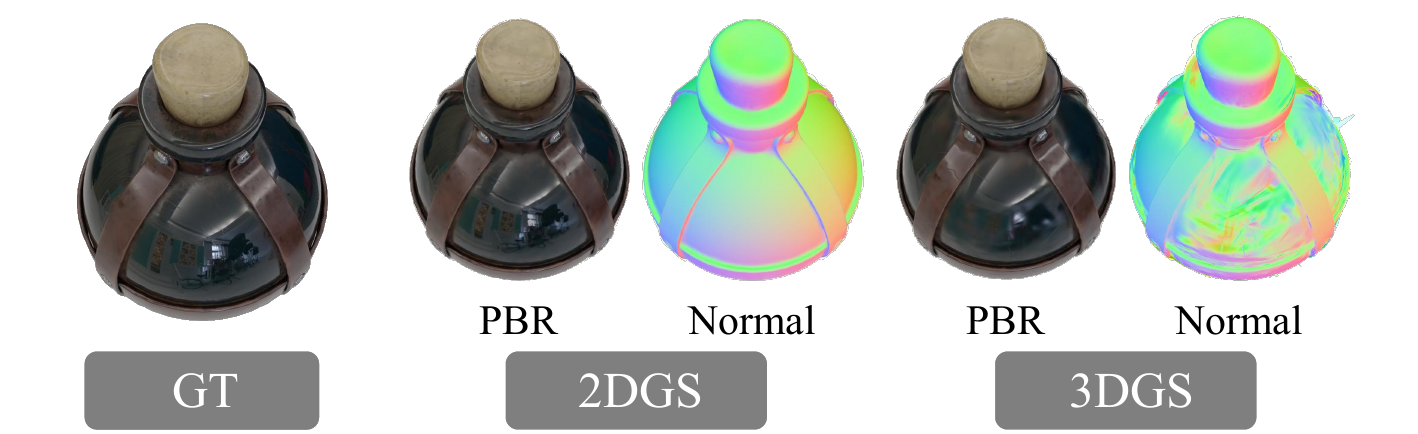}
        \caption{Ablation study on usage of 2DGS.}
        \label{fig:2DGS}
    \end{minipage}
\end{figure}

\begin{figure}[!t]
    \centering
    \begin{minipage}{0.44\linewidth}
        \centering
        \includegraphics[width=\linewidth]{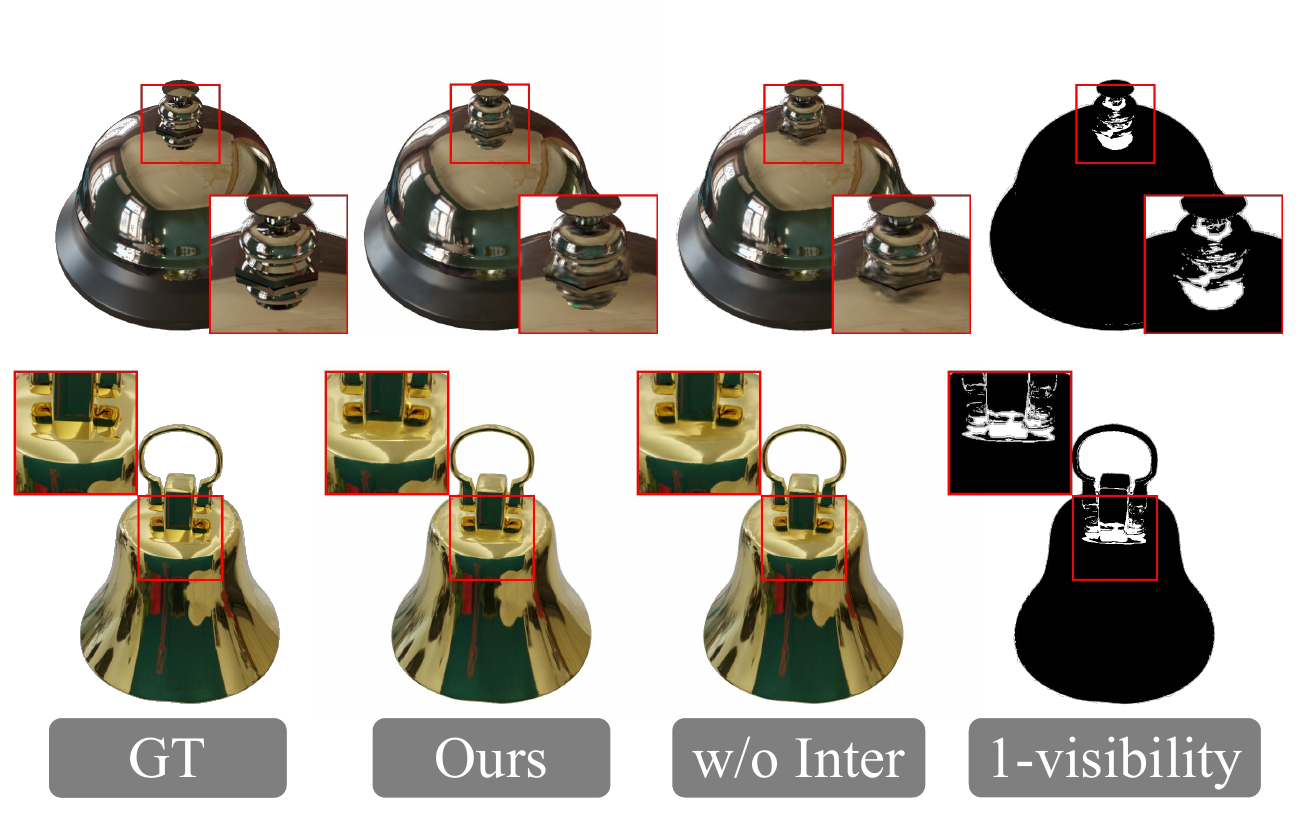}
        \caption{Ablation study on inter-reflection.}
        \label{fig:Indirect}
    \end{minipage}
    \hspace{0.02\linewidth}
    \begin{minipage}{0.52\linewidth}
        \centering
        \includegraphics[width=\linewidth]{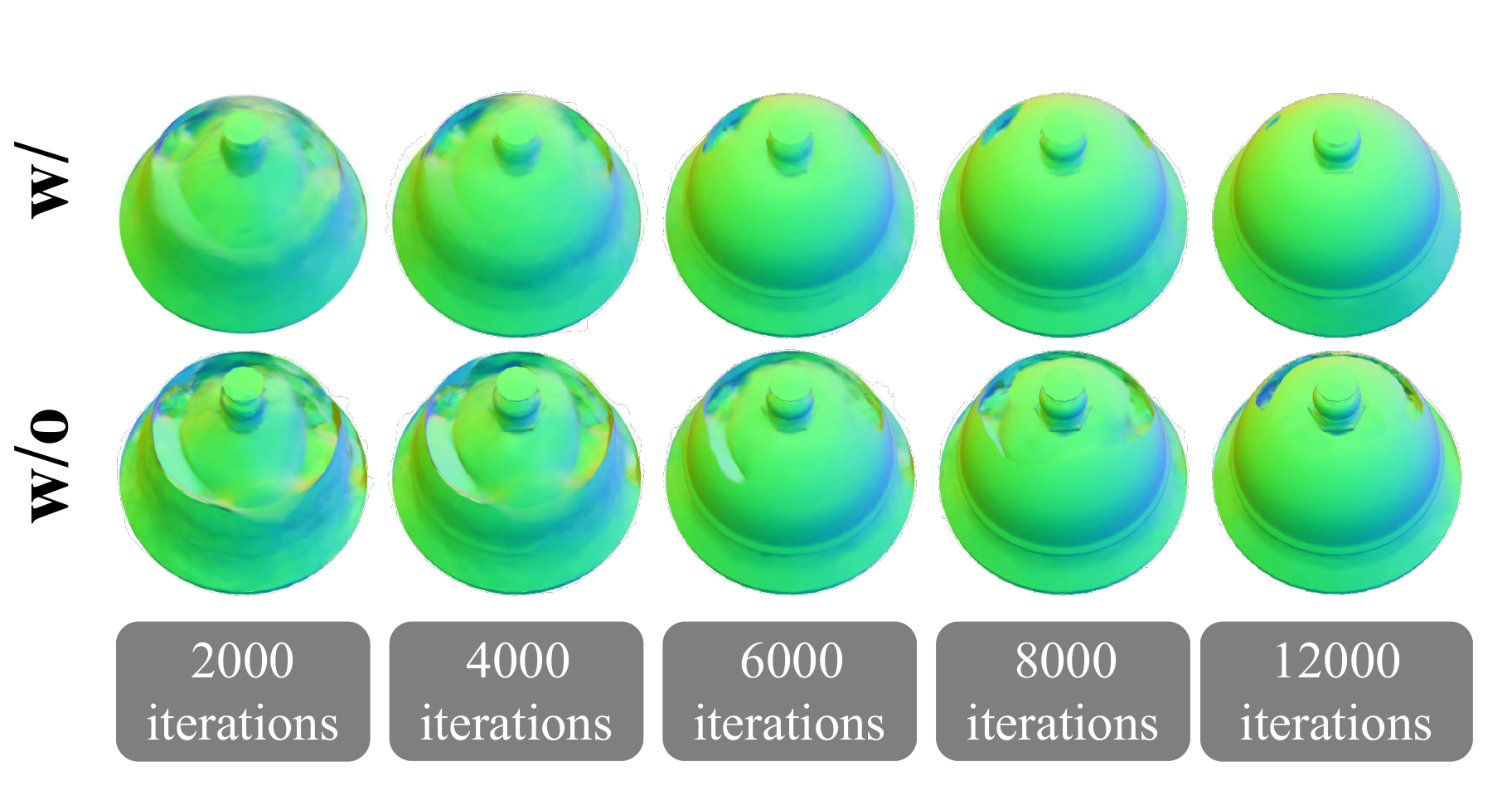}
        \caption{Effect of material-aware normal propagation on normal maps at various steps.}
        \label{fig:Normal prop}
    \end{minipage}
\end{figure}
\begin{table}[t]
    \centering
    \begin{minipage}{0.73\linewidth}
        \centering
        \caption{Quantitative comparison on normal maps, environment maps, and efficiency, with all metrics averaged across datasets.}
        \label{tb:Comprehensive Comparison}
        \vspace{0.3cm} 
        \renewcommand{\arraystretch}{1.2}
        \resizebox{\textwidth}{!}{
        \begin{tabular}{l|ccc|c|cc}
        \hline
        \hline
        \textbf{Model} & \multicolumn{3}{c|}{\textbf{Normal}} & \multicolumn{1}{c|}{\textbf{Envmap}} & \multicolumn{2}{c}{\textbf{Efficiency}}\\
        \hline
         & MAE$\downarrow$ & SSIM$\uparrow$ & LPIPS$\downarrow$ & LPIPS$\downarrow$ & Training time (h)$\downarrow$ & FPS$\uparrow$\\
        \hline
        {ENVIDR}~\citep{liang2023envidr} & 2.74 & \cellcolor{red!10}0.948 & 0.0728 & \cellcolor{red!10}0.5099 & 5.84 & 1 \\
        {GShader}~\citep{jiang2024gaussianshader} & 7.00 & 0.931 & 0.1109 & 0.6145 & \cellcolor{red!10}0.48 & 28 \\
        {3DGS-DR}~\citep{ye20243d} & \cellcolor{red!10}2.62 & 0.947 & \cellcolor{red!10}0.0645 & 0.5290 & \cellcolor{red!50}0.35 & \cellcolor{red!50}251 \\
        \textbf{\model{}} & \cellcolor{red!50}2.15 & \cellcolor{red!50}0.954 & \cellcolor{red!50}0.0511 & \cellcolor{red!50}0.4431 & 0.58 & \cellcolor{red!10}122 \\
        \hline
        \hline
        \end{tabular}
        }
    \end{minipage}
    \hspace{0.02\linewidth}
    \begin{minipage}{0.23\linewidth}
        \centering
        \caption{Ablation studies on geometry optimization techniques.}
        \label{tb:mae_comparison}
        \vspace{0cm} 
        \renewcommand{\arraystretch}{1.2}
        \resizebox{\textwidth}{!}{
        \begin{tabular}{l|c}
        \hline
        \hline
        \textbf{Model} & \textbf{MAE} \\
        \hline
        \textbf{\textbf{\model{}}} & \cellcolor{red!50}2.15 \\
        \textbf{w/o 2DGS} & 4.45 \\
        \textbf{w/o Initial stage} & \cellcolor{red!10}3.53 \\
        \textbf{w/o Material-aware} & 3.86 \\
        \textbf{w/o PBR} & 3.81 \\
        \hline
        \hline
        \end{tabular}
        }
    \end{minipage}
\end{table}
Furthermore, \model{} excels in optimization speed, demonstrating rapid convergence, and supports real-time rendering. The average training time and frame-per-second (FPS) rendering performance are evaluated on the Shiny Blender dataset~\citep{verbin2022ref}.

\subsection{Ablation study}

We provide quantitative ablation studies for \model{} below.

\paragraph{Physically based rendering}
Assigning BRDF properties to Gaussians greatly enhances their ability to represent complex scenes. Compared to the simplified shading function using only one reflective direction vector to query the environment map as in 3DGS-DR~\citep{ye20243d},  \model{} demonstrates substantially better performance when learning scenes with material variations, such as the rough band on the sphere (Figure~\ref{fig:PBR}) and the visor on the helmet (Figure~\ref{fig:compare-main}). Also, physically based rendering contributes to the geometry reconstruction, as verified by the results in Table~\ref{tb:mae_comparison}.

\paragraph{Gaussian-grounded inter-reflection}
As shown in Figure \ref{fig:Indirect},
the impact of our inter-reflection component is particularly evident on the head part of Tbell and Bell. With ray-traced visibility, we manage to reconstruct indirect lighting for novel view synthesis. Note that, inter-reflection is the minority in the glossy synthetic dataset. Its effect cannot be fully observed from Table~\ref{tb:ablation}.

\paragraph{Deferred rendering}
We conduct physically based rendering at the per-Gaussian level and observe that the modeling of reflective details becomes noticeably more blurred. As shown in Table~\ref{tb:ablation}, without deferred rendering, all quality metrics of the rendered output drop significantly.

\begin{table*}[!t]
\centering

\caption{Ablation studies on the components of \model{}. w/o Material-aware: use original normal propagation following 3DGS-DR~\citep{ye20243d}.}
\label{tb:ablation}
\vspace{0.01cm} 
\renewcommand{\arraystretch}{1.2}
\resizebox{1.0\textwidth}{!}{
\begin{tabular}{l|c|c|c|c|c|c}
\hline
\hline
\textbf{Model} & {\textbf{\model{}}} & {\textbf{w/o PBR}} & {\textbf{w/o Inter-reflection}} & {\textbf{w/o Deferred rendering}} & {\textbf{w/o Initial stage}} & {\textbf{w/o Material-aware}} \\
\hline

\textbf{PSNR$\uparrow$} & \cellcolor{red!50}30.33 & 29.84 & \cellcolor{red!10}30.14 & 28.85 & 29.94 & 29.68 \\
\textbf{SSIM$\uparrow$} & \cellcolor{red!50}0.958 & 0.956 & \cellcolor{red!10}0.957 & 0.935 & 0.956 & 0.952 \\
\textbf{LPIPS$\downarrow$} & \cellcolor{red!50}0.049 & 0.052 & \cellcolor{red!10}0.050 &0.074 & 0.052 & 0.056 \\
\hline
\hline
\end{tabular}
}
\end{table*}

\paragraph{2D Gaussian primitive}
The results, as shown in Figure~\ref{fig:2DGS} , Table~\ref{tb:mae_comparison} and Table~\ref{tb:2D_compare} in the appendix, emphasize the superiority of 2D Gaussian primitives over 3D alternative for more precise geometric representation. This is because 3D Gaussian conflicts with the thin nature of surfaces for its volumetric radiance representation. Notably, replacing 2DGS with 3DGS and keeping other components unchanged, Ref-Gaussian can still achieve outstanding performance across Shiny Blender~\citep{verbin2022ref} and Glossy Synthetic~\citep{liu2023nero} datasets, as shown in Table~\ref{tb:2D_compare} in the appendix.

\paragraph{Per-Gaussian shading initialization}
The per-Gaussian physically based rendering used in the initial stage offers notable advantages in both the speed and accuracy of geometric convergence. When we replace the initial stage with an equal number of steps using physically based deferred rendering, the final results show a decline in rendering quality and normal reconstruction quality across various metrics.

\paragraph{Material-aware normal propagation}
The material-aware normal propagation accelerates the convergence of smooth surfaces. The quantitative comparisons presented in Tables~\ref{tb:mae_comparison} and~\ref{tb:ablation} demonstrate its effectiveness in geometry reconstruction. As shown in Figure \ref{fig:Normal prop}, smooth surfaces are prone to geometric collapse, but our material-aware normal propagation effectively addresses this issue. 

\subsection{Application to relighting and editing}
\model{} is a generic model, supporting many applications
such as relighting and editing. 
For relighting, we model the diffuse term using integrated environment lighting,
allowing to be modified.
Similarly, we can modify the material properties of the Gaussians to edit the reconstructed scenes.
We showcase the results in Figure~\ref{fig:compare-relight}~(appendix).

\section{Conclusion}
In this paper, we have introduced \textbf{\model{}}, a novel Gaussian splatting framework designed for reconstruction of reflective objects/scenes in real time and high quality. 
This is achieved by formulating two integrated components:
physically based deferred rendering and Gaussian grounded inter-reflection,
along with geometry focused model optimisation.
We also show the importance of choosing the basis representation model
with 2DGS preferred over 3D alternative. 
Extensive experiments on multiple datasets demonstrate the superiority of \model{} over existing methods in terms of quantitative metrics, visual quality, and efficiency. Also, we showcase the effect of \model{} for other applications such as relighting and editing.

\bibliography{main}

\begin{thebibliography}{34}
\providecommand{\natexlab}[1]{#1}
\providecommand{\url}[1]{\texttt{#1}}
\expandafter\ifx\csname urlstyle\endcsname\relax
  \providecommand{\doi}[1]{doi: #1}\else
  \providecommand{\doi}{doi: \begingroup \urlstyle{rm}\Url}\fi

\bibitem[Barron et~al.(2021)Barron, Mildenhall, Tancik, Hedman, Martin-Brualla, and Srinivasan]{barron2021mip}
J.~T. Barron, B.~Mildenhall, M.~Tancik, P.~Hedman, R.~Martin-Brualla, and P.~P. Srinivasan.
\newblock Mip-nerf: A multiscale representation for anti-aliasing neural radiance fields.
\newblock In \emph{ICCV}, 2021.

\bibitem[Barron et~al.(2022)Barron, Mildenhall, Verbin, Srinivasan, and Hedman]{barron2022mip}
J.~T. Barron, B.~Mildenhall, D.~Verbin, P.~P. Srinivasan, and P.~Hedman.
\newblock Mip-nerf 360: Unbounded anti-aliased neural radiance fields.
\newblock In \emph{CVPR}, 2022.

\bibitem[Barron et~al.(2023)Barron, Mildenhall, Verbin, Srinivasan, and Hedman]{barron2023zipnerf}
J.~T. Barron, B.~Mildenhall, D.~Verbin, P.~P. Srinivasan, and P.~Hedman.
\newblock Zip-nerf: Anti-aliased grid-based neural radiance fields.
\newblock In \emph{ICCV}, 2023.

\bibitem[Boss et~al.(2021{\natexlab{a}})Boss, Braun, Jampani, Barron, Liu, and Lensch]{boss2021nerd}
M.~Boss, R.~Braun, V.~Jampani, J.~T. Barron, C.~Liu, and H.~Lensch.
\newblock Nerd: Neural reflectance decomposition from image collections.
\newblock In \emph{ICCV}, 2021{\natexlab{a}}.

\bibitem[Boss et~al.(2021{\natexlab{b}})Boss, Jampani, Braun, Liu, Barron, and Lensch]{boss2021neural}
M.~Boss, V.~Jampani, R.~Braun, C.~Liu, J.~Barron, and H.~Lensch.
\newblock Neural-pil: Neural pre-integrated lighting for reflectance decomposition.
\newblock \emph{NeurIPS}, 2021{\natexlab{b}}.

\bibitem[Boss et~al.(2022)Boss, Engelhardt, Kar, Li, Sun, Barron, Lensch, and Jampani]{boss2022samurai}
M.~Boss, A.~Engelhardt, A.~Kar, Y.~Li, D.~Sun, J.~Barron, H.~Lensch, and V.~Jampani.
\newblock Samurai: Shape and material from unconstrained real-world arbitrary image collections.
\newblock \emph{NeurIPS}, 2022.

\bibitem[Burley and Studios(2012)]{burley2012physically}
B.~Burley and W.~D.~A. Studios.
\newblock Physically-based shading at disney.
\newblock In \emph{SIGGRAPH}, 2012.

\bibitem[Chen et~al.(2022)Chen, Xu, Geiger, Yu, and Su]{Chen2022ECCV}
A.~Chen, Z.~Xu, A.~Geiger, J.~Yu, and H.~Su.
\newblock Tensorf: Tensorial radiance fields.
\newblock In \emph{ECCV}, 2022.

\bibitem[Fridovich-Keil et~al.(2022)Fridovich-Keil, Yu, Tancik, Chen, Recht, and Kanazawa]{yu_and_fridovichkeil2021plenoxels}
S.~Fridovich-Keil, A.~Yu, M.~Tancik, Q.~Chen, B.~Recht, and A.~Kanazawa.
\newblock Plenoxels: Radiance fields without neural networks.
\newblock In \emph{CVPR}, 2022.

\bibitem[Gao et~al.(2023)Gao, Gu, Lin, Zhu, Cao, Zhang, and Yao]{gao2023relightable}
J.~Gao, C.~Gu, Y.~Lin, H.~Zhu, X.~Cao, L.~Zhang, and Y.~Yao.
\newblock Relightable 3d gaussian: Real-time point cloud relighting with brdf decomposition and ray tracing.
\newblock \emph{arXiv preprint}, 2023.

\bibitem[Huang et~al.(2024)Huang, Yu, Chen, Geiger, and Gao]{huang20242d}
B.~Huang, Z.~Yu, A.~Chen, A.~Geiger, and S.~Gao.
\newblock 2d gaussian splatting for geometrically accurate radiance fields.
\newblock In \emph{SIGGRAPH}, 2024.

\bibitem[Jiang et~al.(2024)Jiang, Tu, Liu, Gao, Long, Wang, and Ma]{jiang2024gaussianshader}
Y.~Jiang, J.~Tu, Y.~Liu, X.~Gao, X.~Long, W.~Wang, and Y.~Ma.
\newblock Gaussianshader: 3d gaussian splatting with shading functions for reflective surfaces.
\newblock In \emph{CVPR}, 2024.

\bibitem[Jin et~al.(2023)Jin, Liu, Xu, Zhang, Han, Bi, Zhou, Xu, and Su]{jin2023tensoir}
H.~Jin, I.~Liu, P.~Xu, X.~Zhang, S.~Han, S.~Bi, X.~Zhou, Z.~Xu, and H.~Su.
\newblock Tensoir: Tensorial inverse rendering.
\newblock In \emph{CVPR}, 2023.

\bibitem[Kerbl et~al.(2023)Kerbl, Kopanas, Leimk{\"u}hler, and Drettakis]{kerbl20233d}
B.~Kerbl, G.~Kopanas, T.~Leimk{\"u}hler, and G.~Drettakis.
\newblock 3d gaussian splatting for real-time radiance field rendering.
\newblock \emph{ACM Trans. Graph.}, 2023.

\bibitem[Liang et~al.(2023)Liang, Chen, Li, Chen, Panneer, and Vijaykumar]{liang2023envidr}
R.~Liang, H.~Chen, C.~Li, F.~Chen, S.~Panneer, and N.~Vijaykumar.
\newblock Envidr: Implicit differentiable renderer with neural environment lighting.
\newblock In \emph{ICCV}, 2023.

\bibitem[Liang et~al.(2024)Liang, Zhang, Feng, Shan, and Jia]{liang2024gs}
Z.~Liang, Q.~Zhang, Y.~Feng, Y.~Shan, and K.~Jia.
\newblock Gs-ir: 3d gaussian splatting for inverse rendering.
\newblock In \emph{CVPR}, 2024.

\bibitem[Liu et~al.(2023)Liu, Wang, Lin, Long, Wang, Liu, Komura, and Wang]{liu2023nero}
Y.~Liu, P.~Wang, C.~Lin, X.~Long, J.~Wang, L.~Liu, T.~Komura, and W.~Wang.
\newblock Nero: Neural geometry and brdf reconstruction of reflective objects from multiview images.
\newblock \emph{ACM Trans. Graph.}, 2023.

\bibitem[Mildenhall et~al.(2020)Mildenhall, Srinivasan, Tancik, Barron, Ramamoorthi, and Ng]{mildenhall2020nerf}
B.~Mildenhall, P.~P. Srinivasan, M.~Tancik, J.~T. Barron, R.~Ramamoorthi, and R.~Ng.
\newblock Nerf: Representing scenes as neural radiance fields for view synthesis.
\newblock In \emph{ECCV}, 2020.

\bibitem[Mildenhall et~al.(2021)Mildenhall, Srinivasan, Tancik, Barron, Ramamoorthi, and Ng]{mildenhall2021nerf}
B.~Mildenhall, P.~P. Srinivasan, M.~Tancik, J.~T. Barron, R.~Ramamoorthi, and R.~Ng.
\newblock Nerf: Representing scenes as neural radiance fields for view synthesis.
\newblock \emph{Commun. ACM}, 2021.

\bibitem[M{\"u}ller et~al.(2022)M{\"u}ller, Evans, Schied, and Keller]{muller2022instant}
T.~M{\"u}ller, A.~Evans, C.~Schied, and A.~Keller.
\newblock Instant neural graphics primitives with a multiresolution hash encoding.
\newblock \emph{TOG}, 2022.

\bibitem[Munkberg et~al.(2022)Munkberg, Hasselgren, Shen, Gao, Chen, Evans, M{\"u}ller, and Fidler]{munkberg2022extracting}
J.~Munkberg, J.~Hasselgren, T.~Shen, J.~Gao, W.~Chen, A.~Evans, T.~M{\"u}ller, and S.~Fidler.
\newblock Extracting triangular 3d models, materials, and lighting from images.
\newblock In \emph{CVPR}, 2022.

\bibitem[Shi et~al.(2023)Shi, Wu, Wu, Liu, Zhao, Feng, Liu, Zhang, Zhang, Zhou, et~al.]{shi2023gir}
Y.~Shi, Y.~Wu, C.~Wu, X.~Liu, C.~Zhao, H.~Feng, J.~Liu, L.~Zhang, J.~Zhang, B.~Zhou, et~al.
\newblock Gir: 3d gaussian inverse rendering for relightable scene factorization.
\newblock \emph{arXiv preprint}, 2023.

\bibitem[Srinivasan et~al.(2021)Srinivasan, Deng, Zhang, Tancik, Mildenhall, and Barron]{srinivasan2021nerv}
P.~P. Srinivasan, B.~Deng, X.~Zhang, M.~Tancik, B.~Mildenhall, and J.~T. Barron.
\newblock Nerv: Neural reflectance and visibility fields for relighting and view synthesis.
\newblock In \emph{CVPR}, 2021.

\bibitem[Sun et~al.(2022)Sun, Sun, and Chen]{SunSC22}
C.~Sun, M.~Sun, and H.~Chen.
\newblock Direct voxel grid optimization: Super-fast convergence for radiance fields reconstruction.
\newblock In \emph{CVPR}, 2022.

\bibitem[Verbin et~al.(2022)Verbin, Hedman, Mildenhall, Zickler, Barron, and Srinivasan]{verbin2022ref}
D.~Verbin, P.~Hedman, B.~Mildenhall, T.~Zickler, J.~T. Barron, and P.~P. Srinivasan.
\newblock Ref-nerf: Structured view-dependent appearance for neural radiance fields.
\newblock In \emph{CVPR}, 2022.

\bibitem[Wang et~al.(2021)Wang, Liu, Liu, Theobalt, Komura, and Wang]{wang2021neus}
P.~Wang, L.~Liu, Y.~Liu, C.~Theobalt, T.~Komura, and W.~Wang.
\newblock Neus: Learning neural implicit surfaces by volume rendering for multi-view reconstruction.
\newblock \emph{arXiv preprint}, 2021.

\bibitem[Wang et~al.(2004)Wang, Bovik, Sheikh, and Simoncelli]{wang2004image}
Z.~Wang, A.~C. Bovik, H.~R. Sheikh, and E.~P. Simoncelli.
\newblock Image quality assessment: from error visibility to structural similarity.
\newblock \emph{IEEE Trans. Image Process.}, 2004.

\bibitem[Wu et~al.(2024)Wu, Bi, Xu, Luan, Zhang, Georgiev, Sunkavalli, and Ramamoorthi]{wu2024neural}
L.~Wu, S.~Bi, Z.~Xu, F.~Luan, K.~Zhang, I.~Georgiev, K.~Sunkavalli, and R.~Ramamoorthi.
\newblock Neural directional encoding for efficient and accurate view-dependent appearance modeling.
\newblock In \emph{CVPR}, 2024.

\bibitem[Yao et~al.(2022)Yao, Zhang, Liu, Qu, Fang, McKinnon, Tsin, and Quan]{yao2022neilf}
Y.~Yao, J.~Zhang, J.~Liu, Y.~Qu, T.~Fang, D.~McKinnon, Y.~Tsin, and L.~Quan.
\newblock Neilf: Neural incident light field for physically-based material estimation.
\newblock In \emph{ECCV}, 2022.

\bibitem[Ye et~al.(2024)Ye, Hou, and Zhou]{ye20243d}
K.~Ye, Q.~Hou, and K.~Zhou.
\newblock 3d gaussian splatting with deferred reflection.
\newblock In \emph{SIGGRAPH}, 2024.

\bibitem[Yu et~al.(2024)Yu, Sattler, and Geiger]{yu2024gaussian}
Z.~Yu, T.~Sattler, and A.~Geiger.
\newblock Gaussian opacity fields: Efficient and compact surface reconstruction in unbounded scenes.
\newblock \emph{arXiv preprint}, 2024.

\bibitem[Zhang et~al.(2023)Zhang, Yao, Li, Liu, Fang, McKinnon, Tsin, and Quan]{zhang2023neilf++}
J.~Zhang, Y.~Yao, S.~Li, J.~Liu, T.~Fang, D.~McKinnon, Y.~Tsin, and L.~Quan.
\newblock Neilf++: Inter-reflectable light fields for geometry and material estimation.
\newblock In \emph{ICCV}, 2023.

\bibitem[Zhang et~al.(2018)Zhang, Isola, Efros, Shechtman, and Wang]{zhang2018unreasonable}
R.~Zhang, P.~Isola, A.~A. Efros, E.~Shechtman, and O.~Wang.
\newblock The unreasonable effectiveness of deep features as a perceptual metric.
\newblock In \emph{CVPR}, 2018.

\bibitem[Zhang et~al.(2021)Zhang, Srinivasan, Deng, Debevec, Freeman, and Barron]{zhang2021nerfactor}
X.~Zhang, P.~P. Srinivasan, B.~Deng, P.~Debevec, W.~T. Freeman, and J.~T. Barron.
\newblock Nerfactor: Neural factorization of shape and reflectance under an unknown illumination.
\newblock \emph{ACM Trans. Graph.}, 2021.

\end{thebibliography}
\bibliographystyle{abbrvnat}

\newpage
\appendix
\section{Appendix}

\subsection{Limitations}
As \model{} is specifically designed for reconstructing reflective scenes, it might not excel for non-specular scenes. When modeling inter-reflections, we only consider the visibility in the reflection direction to approximate the overall visibility across the entire specular lobe. While this approach significantly reduces the computational complexity of rendering, it might introduce some noise in material estimation due to the inherent inaccuracy with the approximation.

\subsection{Qualitative results of relighting and editing}
\begin{figure}[!h]
    \centering
    \vspace{0.01cm} 
    \includegraphics[width=\linewidth]{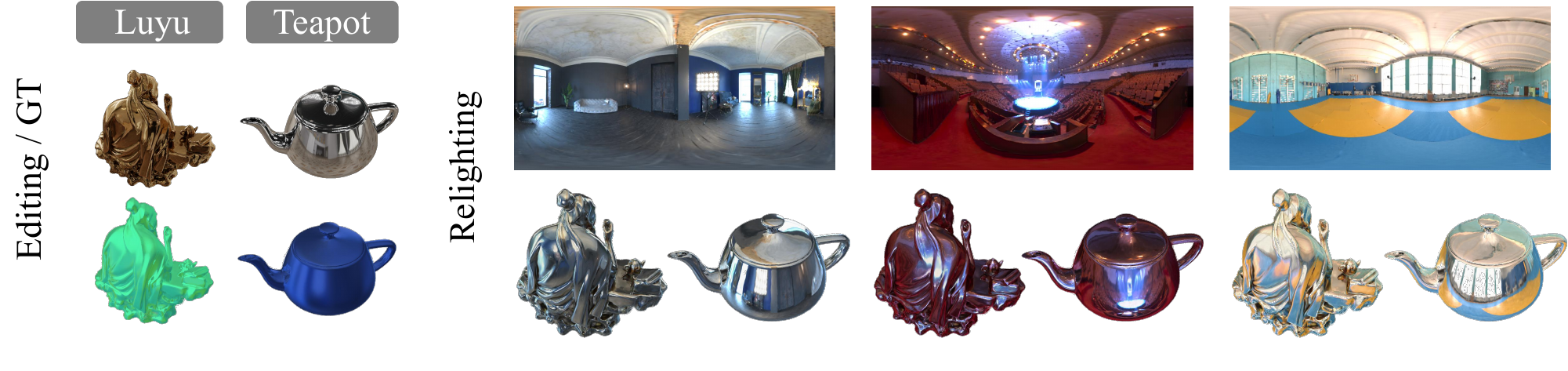}
    \caption{
    Relighting and editing on Luyu and Teapot with ground truth on the left.
    }
    \label{fig:compare-relight}
\end{figure}

We illustrate the results of both applications with realistic visual effects in Figure~\ref{fig:compare-relight}.

\subsection{Further quantitative comparisons on non-specular dataset}

By design, \model{} excel for both reflective and non-reflective scenes, \textbf{making it as a unified soluton for modeling a variety of scenes with varying reflection}. This is because, as described in Eq.\ref{eq:indirect}, inter-reflection is incorporated just as an additional element into the specular component. Consequently, in less reflective scenes, the influence of inter-reflection diminishes naturally and accordingly as the metallic attribute decreases. In such cases, the diffuse component instead takes over, including the capture of any (usually substantially weak) inter-reflection effect to achieve top performance. This behavior is evident in the decomposition results provided in Figure~\ref{fig:inverse_mesh} and Figure~\ref{fig:vis-appendix} in the appendix. In contrast, previous reflection focused alternatives such as 3DGS-DR~\citep{ye20243d} are designed specifically only for reflective scenes, leading to narrowed applications as mentioned.

To support this claim, we have included Table~\ref{tb:nerf_synthetic_compare} in the appendix, which presents a per-scene quantitative comparison among \model{}, 3DGS-DR~\citep{ye20243d}, and 3DGS~\citep{kerbl20233d} on the NeRF-Synthetic dataset~\citep{mildenhall2020nerf}, predominantly composed of non-reflective scenes, for novel view synthesis. The results show that our \model{} still excels over both alternatives, validating its generic and unified advantages.

\subsection{Extra baselines for NeRF-based method}
 We have further compared the average PSNR, SSIM, LPIPS, and FPS on the Glossy Blender dataset with NeRO~\citep{liu2023nero} and NeuS~\citep{wang2021neus} in Table~\ref{tb:neus_nero}, as well as with NDE~\citep{wu2024neural} on the Shiny Blender dataset in Table~\ref{tb:nde}. The quantitative results demonstrate that \model{} significantly outperforms NeRO and remains comparable to NDE in rendering quality while achieving significantly better rendering speed and training efficiency.

\begin{table}[!h]

\centering
\caption{NVS quality of NeuS~\citep{wang2021neus} and NeRO~\citep{liu2023nero} on the Glossy-Blender dataset~\citep{liu2023nero}.}
\vspace{0.1cm}
\label{tb:neus_nero}
\resizebox{0.4\textwidth}{!}{
\begin{tabular}{l|c|c|c}
\hline
\hline
\textbf{Model}  & NeuS & NeRO & {\textbf{\model{}}}\\
\hline
\textbf{PSNR$\uparrow$}  & 27.86 & 29.73 & \cellcolor{red!50}30.33\\
\textbf{SSIM$\uparrow$}  & 0.878 & 0.904 & \cellcolor{red!50}0.958\\
\textbf{LPIPS$\downarrow$}  & 0.375 & 0.324 & \cellcolor{red!50}0.049\\
\textbf{FPS$\uparrow$}  & 0.02 & 0.11 & \cellcolor{red!50}122\\
\hline
\hline
\end{tabular}
}

\end{table}%

\begin{table}[!h]

\centering
\caption{NVS quality of NDE~\citep{wu2024neural} on the Shiny-Blender dataset~\citep{verbin2022ref}.}
\vspace{0.1cm}
\label{tb:nde}
\resizebox{0.35\textwidth}{!}{
\begin{tabular}{l|c|c}
\hline
\hline
\textbf{Model} & NDE & {\textbf{\model{}}}\\
\hline
\textbf{PSNR$\uparrow$} & \cellcolor{red!50}35.48 & 35.04\\
\textbf{SSIM$\uparrow$} & \cellcolor{red!50}0.976 & 0.973 \\
\textbf{LPIPS$\downarrow$} & \cellcolor{red!50}0.027 & 0.056 \\
\textbf{FPS$\uparrow$}  & 66 & \cellcolor{red!50}122 \\
\hline
\hline
\end{tabular}
}
\end{table}

\subsection{Qualitative results of indirect component on Ref-Real Dataset}

\begin{figure}[!h]
    \centering
    \includegraphics[width=\linewidth]{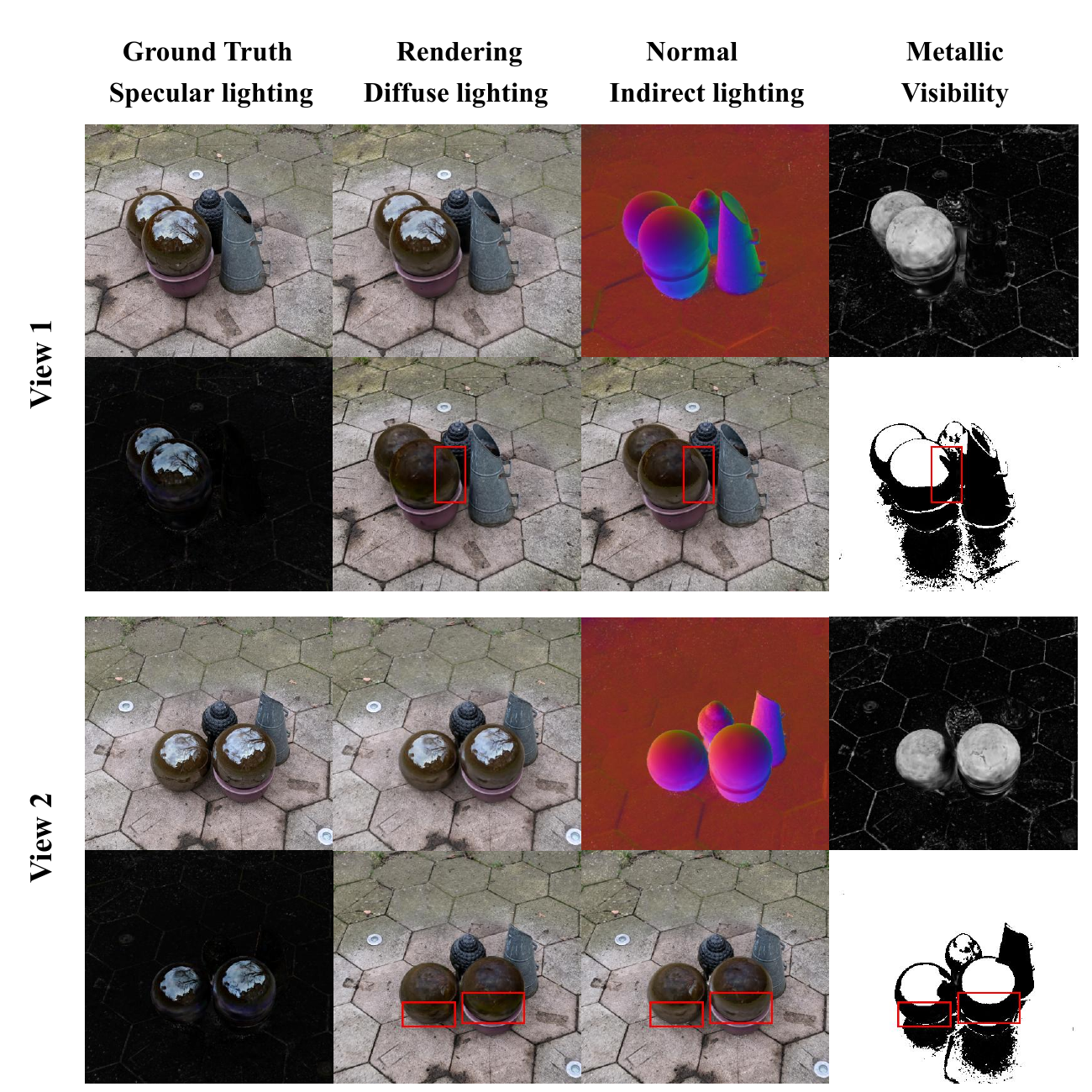}
    \caption{Qualitative results of indirect lighting on Ref-Real dataset~\citep{verbin2022ref}. \textbf{Indirect lighting}: Only considering indirect light as specular component when rendering - a focused view.}
    
    \label{fig:compare-real-ind}
\end{figure}

We additionally provided richer qualitative results in Figure~\ref{fig:compare-real-ind} where multiple objects provide rich inter-reflection. It can be observed that, in the task of novel view synthesis, the three components—diffuse, specular, and indirect light—complement each other effectively and together produce an excellent final rendering result. Specifically, based on the metallic property, the diffuse component primarily models non-reflective objects and ground details outside the stone sphere. The specular component focuses on the parts of the stone sphere where reflected light is not occluded by the ground or other objects, capturing fine reflective details, and the indirect component effectively captures the projections of the ground and adjacent objects. These decomposition results strongly demonstrate that \model{}, even in complex real-world scenes, can accurately capture various details through its well-designed rendering approach and precise geometric reconstruction.

\subsection{Quantitative results of comprehensive 2DGS ablation study}
To explicitly showcase the effect of 2DGS representation and to more fairly and comprehensively demonstrate the superiority of \model{} with other techniques, over other 3DGS-based methods, we replace the 2DGS representation of \model{} with 3DGS while keeping the rest unchanged for comparison as shown in Table~\ref{tb:2D_compare}. The introduction of 2DGS has significantly enhanced the model's performance. And it is worth emphasizing that the \model{} built on 3DGS, also excels in most scenarios of the Shiny Blender~\citep{verbin2022ref} and Glossy Synthetic~\citep{liu2023nero} datasets, further highlighting the inherent superiority of the rendering model.

\subsection{Qualitative results of geometry reconstruction}
\begin{figure}[!h]
    \centering
    \includegraphics[width=\linewidth]{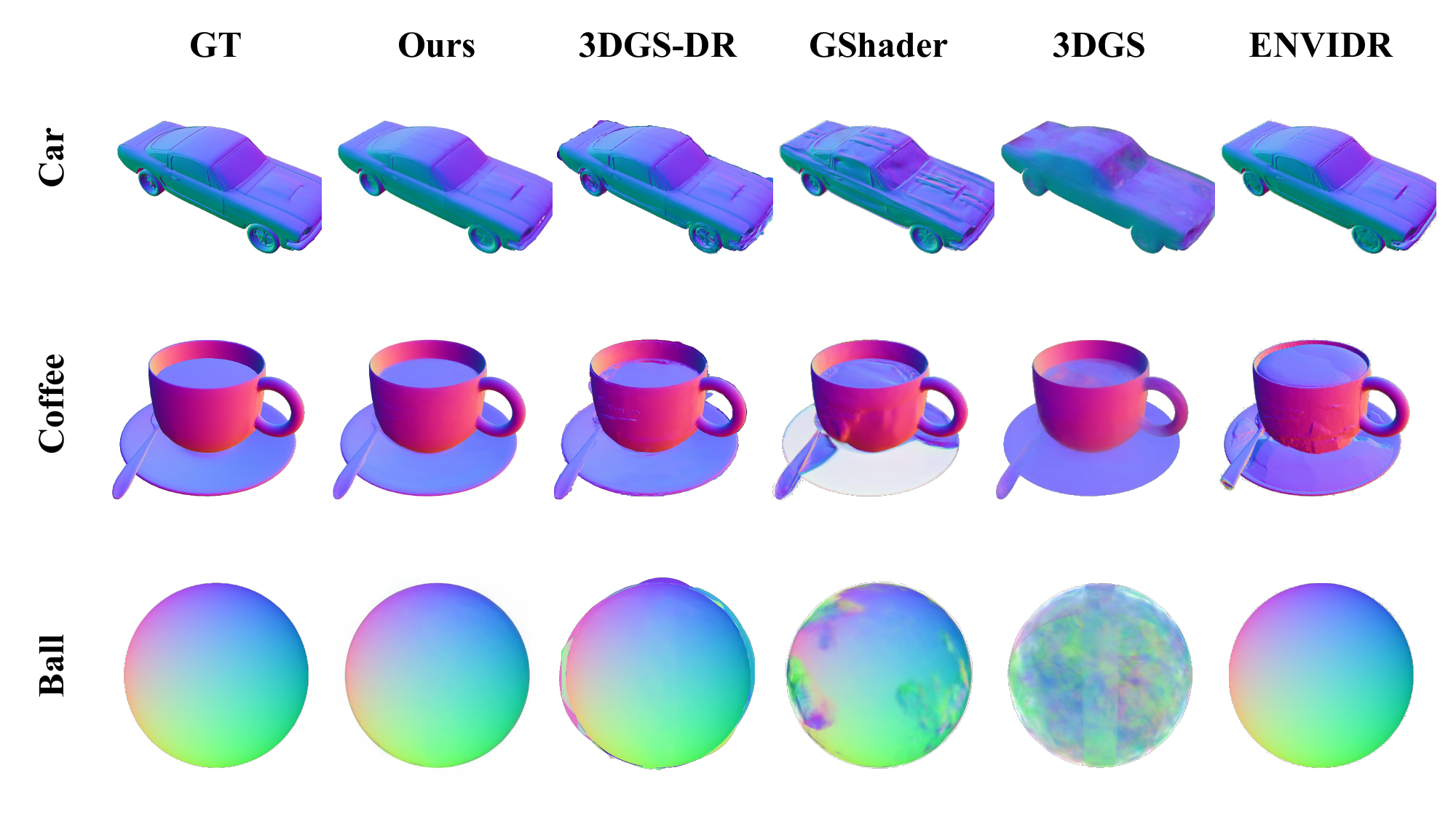}
    \caption{
        Per-scene qualitative comparisons of normals.}
    
    \label{fig:compare-normal}
\end{figure}

\begin{figure}[h]
    \centering
    \includegraphics[width=\linewidth]{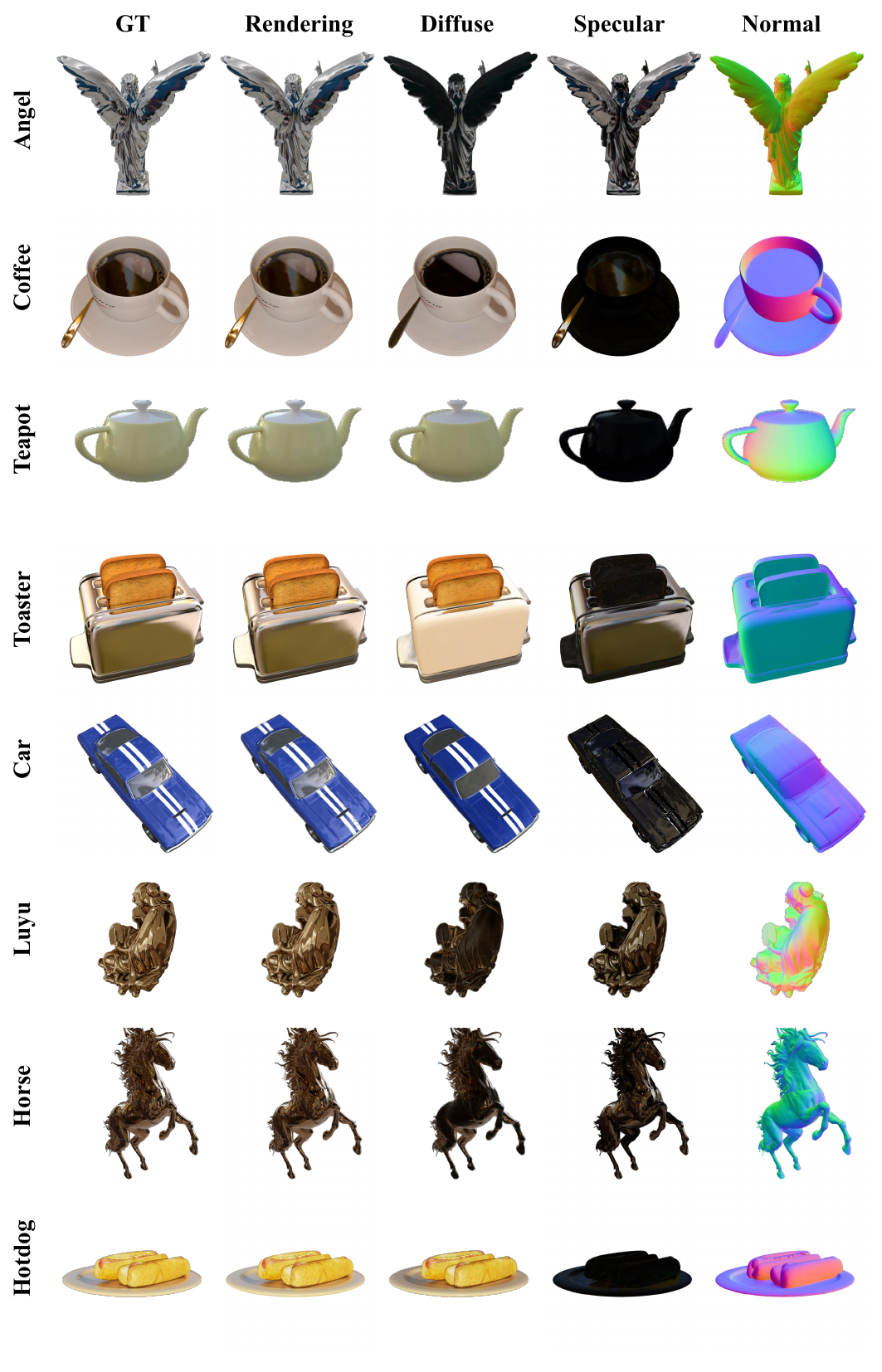}
    \caption{
    Qualitative decomposition results.}
    \label{fig:vis-appendix}
\end{figure}

We provide further illustration of our advantage in geometry reconstruction~(Figure \ref{fig:compare-normal},~\ref{fig:vis-appendix}). The qualitative comparison in Figure~\ref{fig:compare-normal} demonstrates \model{}'s comprehensive grasp of details over other advanced techniques~(such as the tires in car and the water surface in coffee).


\begin{table}[t]
\caption{Per-scene PSNR comparison on synthesized test. w/o 2DGS: Using 3DGS as the representation of our \model{} with the rest unchanged.}
\label{tb:2D_compare}
\vspace{0.1cm} 
\centering
\renewcommand{\arraystretch}{1.3} 
\resizebox{\textwidth}{!}{%
\begin{tabular}{l|cccccc|cccccccc}
\hline
\hline
\textbf{Datasets} & \multicolumn{6}{c|}{\textbf{Shiny Blender \citep{verbin2022ref}}} & \multicolumn{8}{c}{\textbf{Glossy Synthetic \citep{liu2023nero}}} \\
\hline
\textbf{Scenes} & \textbf{ball} & \textbf{car} & \textbf{coffee} & \textbf{helmet} & \textbf{teapot} & \textbf{toaster} & \textbf{angel} & \textbf{bell} & \textbf{cat} & \textbf{horse} & \textbf{luyu} & \textbf{potion} & \textbf{tbell} & \textbf{teapot} \\
\hline
ENVIDR & \cellcolor{red!50}41.02 & 27.81 & 30.57 & 32.71 & 42.62 & 26.03 & 29.02 & \cellcolor{red!10}30.88 & 31.04 & 25.99 & 28.03 & 32.11 & 28.64 & \cellcolor{red!50}26.77\\
3DGS-DR & 33.43 & 30.48 & \cellcolor{red!10}34.53 & 31.44 & \cellcolor{red!10}47.04 & 26.76 & \cellcolor{red!10}29.07 & 30.60 & 32.59 & 26.17 & 28.96 & 32.65 & 29.03 & 25.77\\
\hline
\hline
w/o 2DGS & 36.10 & \cellcolor{red!10}30.65 & 34.51 & \cellcolor{red!50}33.29 & 44.25 & \cellcolor{red!10}27.03 & 28.33 & 30.60 & \cellcolor{red!50}33.14 & \cellcolor{red!10}26.70 & \cellcolor{red!10}29.35 & \cellcolor{red!10}32.94 & \cellcolor{red!10}29.17 & 26.31\\

\textbf{\model{}} & \cellcolor{red!10}37.01 & \cellcolor{red!50}31.04 & \cellcolor{red!50}34.63 & \cellcolor{red!10}32.32 & \cellcolor{red!50}47.16 & \cellcolor{red!50}28.05 & \cellcolor{red!50}30.38 & \cellcolor{red!50}32.86 & \cellcolor{red!10}33.01 & \cellcolor{red!50}27.05 & \cellcolor{red!50}30.04 & \cellcolor{red!50}33.07 & \cellcolor{red!50}29.84 & \cellcolor{red!10}26.68\\

\hline
\hline
\end{tabular}%
}
\end{table}
\begin{table}[t]
    \centering
    \caption{Quantitative comparisons on NeRF-Synthetic dataset~\citep{mildenhall2020nerf}.}
    \renewcommand{\arraystretch}{1.3} 
    \label{tb:nerf_synthetic_compare}
    \resizebox{0.6\textwidth}{!}{%
    \begin{tabular}{c|ccccccc}
        \hline
        \hline
        & \textbf{chair} & \textbf{drums} & \textbf{ficus} & \textbf{hotdog} & \textbf{lego} & \textbf{material} & \textbf{ship} \\
        \hline
        \multicolumn{8}{c}{\textbf{PSNR} $\uparrow$} \\
        \hline
        3DGS & \cellcolor{red!50}{35.03} & \cellcolor{red!10}{26.04} & \cellcolor{red!10}{35.29} & \cellcolor{red!10}{37.57} & \cellcolor{red!50}{33.71} & \cellcolor{red!10}{30.04} & \cellcolor{red!50}{31.43} \\
        
        3DGS-DR & 32.10 & 25.31 & 28.03 & 35.58 & 32.94 & 28.35 & 29.07 \\
        
        \textbf{\model{}} & \cellcolor{red!10}{34.71} & \cellcolor{red!50}{26.33} & \cellcolor{red!50}{35.74} & \cellcolor{red!50}{37.65} & \cellcolor{red!10}{33.46} & \cellcolor{red!50}{30.44} & \cellcolor{red!10}{30.78} \\
        \hline
        \multicolumn{8}{c}{\textbf{SSIM} $\uparrow$} \\
        \hline
        3DGS & \cellcolor{red!50}{0.987} & \cellcolor{red!10}{0.954} & \cellcolor{red!10}{0.987} & \cellcolor{red!50}{0.985} & 0.975 & \cellcolor{red!10}{0.959} & \cellcolor{red!50}{0.906} \\

        3DGS-DR & 0.977 & 0.946 & 0.963 & 0.982 & \cellcolor{red!50}{0.978} & 0.950 & 0.894 \\
        
        \textbf{\model{}} & \cellcolor{red!10}{0.981} & \cellcolor{red!50}{0.955} & \cellcolor{red!50}{0.988} & \cellcolor{red!50}{0.985} & \cellcolor{red!10}{0.976} & \cellcolor{red!50}{0.964} & \cellcolor{red!10}{0.898} \\
        \hline
        \multicolumn{8}{c}{\textbf{LPIPS} $\downarrow$} \\
        \hline
        3DGS & \cellcolor{red!50}{0.012} & 0.\cellcolor{red!10}{040} & \cellcolor{red!50}{0.012} & \cellcolor{red!50}{0.021} & 0.027 & \cellcolor{red!10}{0.040} & \cellcolor{red!50}{0.111} \\

        3DGS-DR & 0.024 & 0.055 & 0.055 & 0.033 & \cellcolor{red!50}{0.026} & 0.044 & 0.129 \\
        
        \textbf{\model{}} & \cellcolor{red!10}{0.021} & \cellcolor{red!50}{0.039} & \cellcolor{red!50}{0.012} & \cellcolor{red!10}{0.025} & \cellcolor{red!50}{0.026} & \cellcolor{red!50}{0.037} & \cellcolor{red!10}{0.125} \\
        \hline
        \hline
    \end{tabular}
    }
\end{table}

\end{document}